\documentclass{article}

\usepackage{geometry}
\usepackage{hyperref}
\usepackage{lmodern}
\usepackage{animate}
\usepackage{tikz}
\usepackage{tikz-cd}
\usetikzlibrary{shapes.misc, positioning}
\usetikzlibrary{arrows,shapes}
\usetikzlibrary{babel}
\usepackage{mathrsfs,amsmath}
\usepackage{mathtools,slashed}

\usepackage{amsthm}
\usepackage{physics}
\usepackage{amsfonts}
\usepackage{amssymb}
\usepackage{graphicx}
\usepackage{wrapfig}
\usepackage{cancel}
\usepackage{braket}
\usepackage[all]{xy}
\usepackage{xcolor}
\usepackage{adjustbox}
\usepackage{subcaption}
\usepackage{capt-of}
\usepackage{tabu}
\makeatletter
\def\fdsy@scale{1.8}
\newcommand\fdsy@mweight@normal{Book}
\newcommand\fdsy@mweight@small{Book}
\newcommand\fdsy@bweight@normal{Medium}
\newcommand\fdsy@bweight@small{Medium}
\DeclareFontFamily{U}{FdSymbolC}{}
\DeclareFontShape{U}{FdSymbolC}{m}{n}{
    <-7.1> s * [\fdsy@scale] FdSymbolC-\fdsy@mweight@small
    <7.1-> s * [\fdsy@scale] FdSymbolC-\fdsy@mweight@normal
}{}
\DeclareFontShape{U}{FdSymbolC}{b}{n}{
    <-7.1> s * [\fdsy@scale] FdSymbolC-\fdsy@bweight@small
    <7.1-> s * [\fdsy@scale] FdSymbolC-\fdsy@bweight@normal
}{}
\makeatother
\DeclareFontFamily{U}{FdSymbolA}{}
\DeclareFontShape{U}{FdSymbolA}{m}{n}{<->FdSymbolA-Book}{}
\DeclareSymbolFont{extrasymbols}{U}{FdSymbolA}{m}{n}
\DeclareMathSymbol{\vardiamondsuit}{\mathord}{extrasymbols}{182}
\DeclareMathSymbol{\varheartsuit}{\mathord}{extrasymbols}{184}
\definecolor{color1}{RGB}{12,18,155}
\definecolor{color2}{RGB}{83,89,155}
\definecolor{color3}{RGB}{75,80,85}

\begin{document}

\title{\textbf{Bases of Steerable Kernels for Equivariant CNNs: From 2D Rotations to the Lorentz Group}}

\author{Alan Garbarz$^\diamondsuit{}^\perp$}
\date{}
\maketitle

\vspace{.25cm}

\begin{minipage}{.9\textwidth}\small \it 
	\begin{center}
    $^\diamondsuit$  Universidad de Buenos Aires, Facultad de Ciencias Exactas y Naturales, Departamento de Física. Ciudad Universitaria, pabell\'on 1, 1428, Buenos Aires, Argentina.
     \end{center}
\end{minipage}

\vspace{.25cm}

\begin{minipage}{.9\textwidth}\small \it 
	\begin{center}
    $^\perp$ CONICET - Universidad de Buenos Aires, Instituto de Física de Buenos Aires (IFIBA). Ciudad Universitaria, pabell\'on 1, 1428, Buenos Aires, Argentina.
     \end{center}
\end{minipage}
\vspace{.5cm}

\maketitle

\begin{abstract}
    We present an alternative way of solving the steerable kernel constraint that appears in the design of steerable equivariant convolutional neural networks. We find explicit  real and complex  bases which are ready to use, for different symmetry groups and for feature maps of arbitrary tensor type. A major advantage of this method is that it bypasses the need to numerically or analytically compute Clebsch-Gordan coefficients and works directly with the representations of the input and output feature maps. The strategy is to find a basis of kernels that respect a simpler invariance condition at some point $x_0$, and then \textit{steer} it with the defining equation of steerability to move to some arbitrary point $x=g\cdot x_0$. This idea has already been mentioned in the literature before, but not advanced in depth and with some generality. Here we describe how it works with minimal technical tools to make it accessible for a general audience.       
    \begin{flushleft}
\hrulefill\\
\footnotesize
{E-mails:  alan@df.uba.ar}
\end{flushleft}
\end{abstract}

\pagebreak

\tableofcontents

\pagebreak

\section{Introduction}

In the last past years it has become clear that incorporating symmetry priors to the design of Convolutional Neural Networks (CNNs) improves considerably its performance. A symmetry hard-wired into a CNN means that if the input is transformed in certain way, the output transforms accordingly, by design. For example, for the task of predicting atomic forces inside a molecule, if the molecule is now rotated arbitrarily,  vectors describing such forces on each atom should rotate as well. In particular, each component of each 3D vector needs to change in a specific way, and a vanilla CNN will have a hard time learning this. On the contrary,  a \textit{rotation equivariant} CNN with the ability to incorporate these rotations does not need to be trained over rotated molecules and can also account for the vector structure on each layer, so the vector components behave as they should under rotations.

In more precise terms, say that  
the transformations $g$ form a group $G$ and this group  acts \footnote{We are considering here what are called group actions $\alpha(g):\text{Feature maps} \rightarrow \text{Feature maps} $, which preserve the group structure. This means $\alpha(g_1)\cdot \alpha(g_2)\cdot=\alpha(g_1g_2)\cdot$.} on the feature maps $F_{\text{in/out}}$:
\begin{equation}
    \alpha_{\text{in}}(g) \cdot F_{\text{in}}=F_{\text{in}}^g,
\end{equation}
and similarly for $F_{\text{out}}$. Now, a layer $L$, with $F_{\text{out}}=L(F_{\text{in}})$, is equivariant with respect to $G$ if
\begin{equation}
   F_{\text{out}}^g=L\left(F^g_{\text{in}}\right),\qquad \forall g\in G
\end{equation}
In the example above of the interatomic forces, $F_{\text{in}}$ could be the initial input, that is the locations and types of atoms, and $F_{\text{out}}$ a vector field of forces, $L$ would be the entire Network and $G$ the group of 3D rotations. 

A CNN is already translation equivariant. The challenge is to augment the translational symmetry to a more general group made of translations and other $G$. There are many works on the design of equivariant CNNs and here we are interested in the so-called steerable equivariant CNNs \cite{cohen2016steerablecnns}.  This is a general approach where feature maps are functions that transform in a particular representation w.r.t. the symmetry group $G$. We give now a more precise description of steerable equivariant CNNs and the steerability constraint on the kernel used to perform the convolution.

\subsection{The steerability constraint}

We have the (Euclidean) input and output feature maps, $F_{\text{in/out}}$, of certain layer of the neural network. More precisely, we think of them as
\begin{equation}
    F_{\text{in/out}}\in L^2(\mathbb{R}^d,\mathbb{R}^{N_{\text{in/out}}})=\left\{F:\mathbb{R}^d\rightarrow \mathbb{R}^{N_{\text{in/out}}};\quad \int_{\mathbb{R}^d}dx\, |F(x)|^{2}<\infty\right\}
\end{equation}
More over, they transform under the affine group\footnote{The affine group Aff$(G)$ is understood here as the semidirect product $(\mathbb{R}^d,+) \rtimes  G $ of translations and some $G \leq$GL$(d)$ acting on $\mathbb{R}^d$.} Aff$(G)$ with the induced representation by $\rho_{\text{in/out}}$, 
\begin{equation}
    \left((t,g)\cdot F_{\text{in/out}}\right)(x)=\rho_{\text{in/out}}(g)F_{\text{in/out}}((t,g)^{-1}\cdot x), \qquad t \in \mathbb{R}^d,\quad g\in G
    .
\end{equation}
Here the $\cdot$ on the LHS is the action of Aff$(G)$ on the space of feature maps (defined by the RHS), where the $\cdot$ on the RHS means its  action on the space $\mathbb{R}^d$, given by $(t,g)^{-1}\cdot x=g^{-1}(x-t) $. These actions are different, but in order to avoid too much notation, we expect it is understood from the context what action $\cdot$ represents. 

A linear layer $L_K$ then maps an input feature map $F_\text{in}$ to an output feature map $F_\text{out}$. A layer $L_K$ is steerable equivariant (or more precisely steerable Aff$(G)$-equivariant ) if  it commutes with the action of the Affine group, namely,
\begin{equation}\label{equivariance}
    (t,g)\cdot (L_K F) = L_K\left((t,g)\cdot F\right)
\end{equation}
It is easy to show that such equivariant condition holds if and only if the layer $L_K$ is a convolution by a kernel $K$, which has to satisfy a steerability constraint. First of all let us state what $K$ is:
\begin{equation}\label{kernel}
    K: \mathbb{R}^d\rightarrow \mathbb{R}^{N_{\text{out}}\times N_{\text{in}}}
\end{equation}
such that the convolution with an input feature map gives something meaningful:
\begin{equation}\label{convolution}
   F_{\text{out}}(x)= \int_{\mathbb{R}^d}dy\, K(x-y)F_{\text{in}}(y)\in L^{2}\left(\mathbb{R}^d,\mathbb{R}^{N_{\text{out}}}\right)
\end{equation}
From this expression and \eqref{equivariance} we get the steerability constraint
\begin{equation}\label{steerabilityconstraint}
    K(g\cdot x)=\frac{1}{|\text{det}\, g|}\rho_{\text{out}}(g)K(x)\rho_{\text{in}}(g)^{-1},\quad \forall\, g\in G
\end{equation}
This steerability constraint and its space of solutions (for different groups $G$) is the main topic of this work. Note that it can be rephrased as saying that $K$ is \textit{invariant under $G$}, since replacing $x\mapsto g^{-1}\cdot x$ we get 
\begin{equation}\label{steerabilityconstraint}
    K(x)=\frac{1}{|\text{det}\, g|}\rho_{\text{out}}(g)K(g^{-1}\cdot x)\rho_{\text{in}}(g)^{-1},\quad \forall\, g\in G
\end{equation}
and the RHS is actually a representation of $G$ on the fields of matrices, labeled by representations $\rho_{\text{out}}$ and $\rho_{\text{in}}$. Therefore, this is saying $K$ remains unchanged under this representation.

When working with compact groups\footnote{When the group is not compact, such as the Lorentz group, we will assume the feature maps of interest are direct sums of irreps, but not unitary irreps.} we can take advantage of the Peter-Weyl Theorem and, by means of a change of basis, decompose each matrix $\rho_{\text{in/out}}$ in a direct sum of unitary irreducible representations $\rho_{j_{\text{in}}}$ and  $\rho_{j_{\text{out}}}$ respectively, where $j_{\text{in}}$ and $j_{\text{out}}$ are indices that label the different unirreps. In this case \eqref{steerabilityconstraint} can be studied for each pair $(j_{\text{in}},j_{\text{out}})$: 
\begin{equation}
    K^{(j_{\text{in}},j_{\text{out}})}(g\cdot x)=\frac{1}{|\text{det}\, g|}\rho_{j_\text{out}}(g)K^{(j_{\text{in}},j_{\text{out}})}(x)\rho_{j_\text{in}}(g)^{-1},\quad \forall\, g\in G\nonumber
.\end{equation}
In order to simplify a little bit the notation, from now on we shall call
 \begin{equation}
     j=j_{\text{out}},\quad l=j_{\text{in}}
 \end{equation}
and then we have 
\begin{equation}\label{steerabilityconstraintjl}
    K^{(j,l)}(g\cdot x)=\frac{1}{|\text{det}\, g|}\rho_{j}(g)K^{(j,l)}(x)\rho_{l}(g)^{-1},\quad \forall\, g\in G
.\end{equation}
The goal is to give a basis of solutions of this (linear) equation for any compact group $G$ and any  allowed pair $(j,l)$. Actually, this has already been accomplished in \cite{lang2021a} using 1) generalized reduced matrix elements, 2) Clebsch-Gordan coefficients, and 3) harmonic basis functions on homogeneous spaces. We are going to show that there is a better approach to obtain the full space of solutions which does not need \textit{a priori} any of this, and harmonic basis functions appear naturally from the matrix elements of $\rho_j$ and $\rho_l$ \footnote{We are not going to use the term ``harmonic'' since it is commonly used for solutions of the Laplace equation, which may be relevant for orthogonal groups but not for other groups.}, and in this way bypassing the need to compute the first two objects. In particular computing Clebsch-Gordan coefficients for some group $G$ can be hard computationally (see \cite{de2006computation} and references therein).  We are going to consider also the (proper orthochronous) Lorentz group SO${}^+(1,3)$, relevant for relativistic systems, which is not compact.

\subsection{Related work}

The first work to put forward an architecture that increased  from translations to another group the equivariance in a CNN was \cite{cohen2016gcnn}. A move towards steerable equivariant CNNs was described in \cite{cohen2016steerablecnns}. The first works implementing rotational equivariant CNNs using steerable kernels were \cite{worrall2017harmonic} and \cite{weiler2018learning}. In order to describe steerable equivariant CNNs in a mathematically precise way, in \cite{cohen2019homogeneous} was shown that the space of steerable kernels on homogeneous spaces can be understood in three different ways, the third being related to the one we are considering here. However we are not going to take double cosets, as in \cite{cohen2019homogeneous}, since it does not seem to be truly necessary. A basis of solutions to \eqref{steerabilityconstraintjl} for 2D rotations and inversions was found in \cite{weiler2019E2CNN}, while for 3D a general method to compute a basis of steerable kernels was proposed in \cite{weiler20183d}. For non-compact groups there are much fewer works, and one relevant to this article is \cite{bogatskiy2020lorentz}, where they designed a steerable equivariant network for the Lorentz group, that is SO$(1,3)$. In that reference the strategy is again that of computing Clebsch-Gordan coefficients, this time for the Lorentz group. As previously explained, we are going to follow a different route, and at the end we will find a basis for Lorentz-steerable kernels in (completely reducible) representations of spacetime tensors, which seems to be the most adequate for relativistic physics.  

A general strategy to fully solve \eqref{steerabilityconstraintjl}  for a compact group on some homogeneous $G-$space was presented in \cite{lang2021a}, while the extension to any $G-$space was later found in \cite{cesa2022program}. As already mentioned,  these works make use of Clebsch-Gordan coefficients for the symmetry group of interest. There are different approaches to solve \eqref{steerabilityconstraintjl} which do not follow the same strategy of the previous references: in \cite{xu2022fourier}  Fourier analysis is used, while \cite{zhdanov2023implicit} and \cite{zhdanov2024CS-CNNs} considered an implicit parametrization of the steerable kernel by MLPs (the latter for the non-compact group $E(p,q)$ which includes the Lorentz group), and in the same direction a practical algorithm for arbitrary matrix groups had been developed in \cite{pmlr-v139-finzi21a}.

\section{A simple solution}\label{simplesolution}

The main insight to fully solve \eqref{steerabilityconstraintjl} is to transform it into a simpler condition on $K$ by using some smaller subgroup $H$ and a fixed point $x_0$, so $K(x_0)$ is in some sense \textit{invariant} under $H$, or put differently $K(x_0)$ is an homomorphism of representations of $H$. Such possible homomorphisms are usually easy to obtain. Having the most general $K(x_0)$, one applies $\rho_j(g)$ to the left and    $\rho_l(g^{-1})$ to the right of $K(x_0)$ as in \eqref{steerabilityconstraintjl} and obtains $K(g \cdot x_0)=K(x)$. This operation is what we call \textit{steering} and the strategy just described has been essentially mentioned before, for instance in \cite{cohen2019homogeneous}.

Let us be more precise: we want to construct a map $K:\mathbb{R}^d \rightarrow \text{Hom}(V_l,V_j)$ that satisfies \eqref{steerabilityconstraintjl}, assuming from now on $|\text{det }g|=1$. As already noted in the literature, \eqref{steerabilityconstraintjl} relates points that belong to the same $G-$orbit, so it is a condition on orbits. More over, for a given $x_0$ on some orbit $G_{x_0}$, one can \textit{define} $K(g\cdot x_0)$ as the RHS of \eqref{steerabilityconstraintjl} evaluated at $x_0$. Any point $x$ in the orbit $G_{x_0}$ is of the form $x=g\cdot x_0$ so we only have to pick the right $g$ to obtain $K(x)$ from $K(x_0)$. However, it could happen that there is another $g'\in G$ such that $g'\cdot x_0=x=g\cdot x_0$, so this means that $g^{-1}g'\in\text{Stab}_{x_0}$, where
\begin{equation}
    \text{Stab}_{x_0}=\left\{g\in G\,;\quad g\cdot x_0=x_0 \right\}
\end{equation}
From now on we shall fix $x_0$ and call $H$ its stabilizer (or equivalently isotropy group):
\begin{equation}
    H:=\text{Stab}_{x_0}
\end{equation}
For consistency, we need then that the RHS of \eqref{steerabilityconstraintjl} gives the same result $K(x_0) $ when evaluated on any $h\in H$:
\begin{equation}\label{constraintisotropy}
    K^{(j,l)}(x_0)=\rho_j(h)K^{(j,l)}(x_0)\rho_l(h^{-1})
\end{equation}
This equation says that $K^{(j,l)}(x_0)$ is an homomorphism  between representations. More precisely, it is an homomorphism between the restriction of representations $\rho_{j/l}$ to $H$. Let us call these restrictions $\rho^H_j$ and $\rho^H_l$, which  are \textit{reducible representations}
of $H$, despite the fact that $\rho_j$ and $\rho_l$ are irreducible representations of $G$. In short, $K^{(j,l)}(x_0)$ is an homomorphism from $\rho_l^H$ to $\rho_j^H$, and is not in general one-to-one.  We have to find the solution space to \eqref{constraintisotropy}, that is a basis of homomorphisms Hom${_H}(V_l,V_j)$, and then \textit{steer}   $K^{(j,l)}(x_0)$ using the RHS of \eqref{steerabilityconstraintjl} with every $g\in G/H$ to obtain $K^{(j,l)}(x)$ with arbitrary $x\in G_{x_0}$.

Notice that compared to \cite{weiler20183d}, we do not need to change to another basis where the tensor product  $\rho_j \otimes \rho_l$ decomposes as direct sum of irreps (the so called coupled basis), and then change back to the original uncoupled basis at the end. This change of bases requires knowing the matrix $Q$ that implements the change of basis, and $Q$ can be written with Clebsch-Gordan coefficients or found by a numerical approach as in \cite{weiler20183d}. In the basis we use, all these steps are unnecessary, and at the end of the day one arrives to the same vector space of solutions, but written with different basis elements.

Let us give a short example with $G=$   SO$(2)$ and orbit given by the unit circle. We can choose $x_0=(1,0)$ and the stabilizer  group  $H$ is given by the identity matrix $\mathbb{I}_{2\times 2}$. Then for real representations labeled by $l,j\geq 1 $  we have $\rho_l(\mathbb{I}_{2\times 2})=\rho_l^H=\mathbb{I}_{2\times 2}$, and the same for $j$. So \eqref{constraintisotropy} is trivial and any $2\times 2$ matrix solves it, therefore we have a basis of 4 elements. When we steer them with the RHS of \eqref{steerabilityconstraintjl}, using that $\rho_l(g)=g^l$, we get the space of solutions of Table 8 of \cite{weiler2019E2CNN}. The same story goes through if either or both representations are trivial: the condition \eqref{constraintisotropy} on $K(x_0)$ is again trivial and we obtain a 1-dimensional vector space of solutions if $j=l=0$ and a 2-dimensional space of solutions if only one of them is the trivial representation. In comparison to the method of \cite{weiler2019E2CNN}, this is a much direct way to solve the constraint \eqref{steerabilityconstraintjl}. We will give more details about this example and others in the following section.    

The problem with this approach is that we have to find homomorphisms between reducible representations of $H$. However, let us note that with \textit{completely reducible} representations, namely those which can be written as a direct sum of irreducible representations, the homomorphisms can be put in very simple form\footnote{When $G$ is semisimple, meaning that is a connected Lie group so that its only closed connected abelian normal subgroup is the trivial subgroup, then any finite-dimensional representation is completely reducible \cite{hall2003lie}. So if $H \leq G$ is a connected subgroup, it is semisimple and any finite-dimensional $H-$ representation is a direct sum of irreps of $H$. Notice that $G$ does not need to be semisimple for our purposes, it is enough for the isotropy group $H$ to be semisimple.}. Let us denote the corresponding decompositions as
\begin{equation}\label{decomposition}
    V_l=\bigoplus_A    V_{l,A},\quad V_j=\bigoplus_B    V_{j,B}
\end{equation}
where $V_{l,A}$ and $V_{j,B}$ are irreducible representations of some group $H$. Then, if $T: V_l\rightarrow V_j$ is an homomorphism of representations, it can be decomposed in blocks
\begin{equation}
    T_{AB}:V_{l,A}\rightarrow V_{j,B}
\end{equation}
and we can write $T$ as a block matrix made of these $T_{A,B}$. Since $T_{A,B}$ are homomorphisms of irreducible representations\footnote{They are intertwiners since we can think of them as the composition $V_{l,A} \hookrightarrow V_l \overset{T}{\rightarrow} V_j \overset{projection}{\longrightarrow} V_{j,B} $ where every map commutes with the corresponding representations of group elements.} we can use Schur's Lemma: we have that $T_{A,B}=0$ if $V_{l,A}$ and $V_{j,B}$ are not isomorphic, and if $V_{l,A}\simeq V_{j,B}$ then $T_{A,B}$ is a scalar multiple of the identity map (the scalar here is a complex number for complex representations) or some other well-known isomorphism (which we will describe in the examples below). We will see how all this works out in specific examples next.

\section{Examples}\label{Examples}

\subsection{SO$(2)$}
 First of all, this is the group of 2D rotations
 \begin{equation}
     g_\phi=\left(\begin{matrix}
    \cos \phi & -\sin \phi\\
    \sin  \phi & \cos \phi
\end{matrix}\right), \qquad \phi\in[0,2\pi)
 \end{equation}
 and its orbits are the origin and the circles. From now on, we work on some circle of arbitrary radius $R$. The isotropy group of any point $x_0$ is made of the identity matrix $\mathbb{I}_{2\times 2}$ only.

\subsubsection*{Complex representations}

The complex unitary irreducible representations are labeled by an integer $n\in \mathbb{Z}$ and are one-dimensional,
\begin{equation}\label{SO(2)unirrepscomplex}
    \rho_n(g_\phi)=e^{in\phi},\qquad n\in \mathbb{Z}
\end{equation}
This case was discussed for real representations previously. We give here its complex version, which is even easier to solve.  

Consider any pair $(j,l)\in \mathbb{Z} \times \mathbb{Z}$, then the constraint \eqref{constraintisotropy} reads 
$$K^{(j,l)}(x_0)=1\times K^{(j,l)}(x_0)\times 1 $$
which is solved by any complex number. Therefore a basis of solutions of the stabilizer constraint is just the number $1$. Now we steer it with \eqref{steerabilityconstraintjl},
$$K^{(j,l)}(g_\phi\cdot x_0)= e^{ij\phi}\times 1 \times  e^{-il\phi}=e^{i(j-l)\phi}$$
and obtain a basis of solutions of \eqref{steerabilityconstraintjl}, \begin{equation}\label{SO(2)solutionljcomplex}
    \mathcal{B}^{(j,l)}_{\text{SO}(2)}=\left\{ e^{i(j-l)\phi}\right\},\qquad \text{(complex representation).}
\end{equation}

\subsubsection*{Real representations}

 The real unirreps of SO$(2)$ are one and two-dimensional representations,
 \begin{equation}\label{SO(2)irreps}
     \rho_0(g_\phi)=1,\quad \rho_j(g_\phi)=\left(\begin{matrix}
    \cos j\phi & -\sin j\phi\\
    \sin j \phi & \cos j\phi
\end{matrix}\right)=e^{j\phi \left(\begin{matrix}
         0 & -1\\
     1 & 0
     \end{matrix}\right)}
 \end{equation}

This case was discussed briefly in the previous Section. Les us give some more detail, but the reader will see there is not much more to say since the method applied to SO$(2)$  immediately gives the desired solutions. Recall what we did before: we picked a point $x_0$ on the orbit, and noticed the stabilizer $H$ is just the 2-by-2 identity matrix, so we have 
\begin{equation}
 \rho^H_0(\mathbb{I}_{2\times 2})=1,\quad 
    \rho_j^H(\mathbb{I}_{2\times 2})=\mathbb{I}_{2\times 2}.
    \end{equation}
Let us work out the three possible cases:
\begin{enumerate}
    \item[$\mathbf{(0,0)}$:] In this case $K^{(0,0)}(x_0)$ is just a real number and \eqref{constraintisotropy} reads $K^{(0,0)}(x_0)=1.K^{(0,0)}(x_0).1$, which is trivial. The solution is just the the real numbers. Steering with the RHS of \eqref{steerabilityconstraintjl} makes no change since both representations $\rho_l=\rho_0$ and $\rho_j=\rho_0$  are trivial and so $K^{(0,0)}(x)=K^{(0,0)}(x_0)$. This means the vector space of solutions is one-dimensional and a basis of solutions can be the number $1$.

    \item[$\mathbf{(0,l)}$:] In this case $K^{(0,l)}(x_0)$ is a 2-component row vector and \eqref{constraintisotropy} reads $K^{(0,l)}(x_0)=1.K^{(0,l)}(x_0).\mathbb{I}_{2\times 2}$, which is again trivial. We can pick a basis of solutions for $K^{(0,l)}(x_0)$ made of the canonical vectors $e_1=(1,0)$ and $e_2=(0,1)$. We then  steer them,
    \begin{align}
        K^{(0,l)}_{e_1}(g_\phi\cdot x_0)&=1.e_1. \rho_l(g_\phi^{-1})=\left(\cos l\phi,\, \sin l\phi \right)\\
        K^{(0,l)}_{e_2}(g_\phi\cdot x_0)&=1.e_2. \rho_l(g_\phi^{-1})=\left(-\sin l\phi,\, \cos l\phi \right)
    \end{align}
and arrive to a basis of solutions to \eqref{steerabilityconstraintjl}:
\begin{equation}\label{SO(2)solution0l}
    \mathcal{B}^{(0,l)}_{\text{SO}(2)}=\left\{ (\cos l\phi,\, \sin l\phi),\,(-\sin l\phi,\, \cos l\phi)  \right\}
\end{equation}
Note that this means that $K^{(0,l)}$ is precisely an arbitrary vector field which is \textit{invariant under $G$} since it is a linear combination of $\hat{r}$ and $\hat{\phi}$. The case $(j,0)$ is completely analogous and can be obtained from $(0,l)$ by taking transpose. 

    \item[$\mathbf{(j,l)}$:] In this case $K^{(j,l)}(x_0)$ is a 2-by-2 matrix and \eqref{constraintisotropy} reads $K^{(j,l)}(x_0)=\mathbb{I}_{2\times 2}.K^{(0,l)}(x_0).\mathbb{I}_{2\times 2}$, which is once again trivial. We can pick a basis of solutions for $K^{(0,l)}(x_0)$ made of the canonical matrices or for example made from the identity and $\sigma_1$, $i\sigma_2$ and $\sigma_3$, where $\sigma_k$ are the Pauli matrices. If we steer for example the matrix $\left(\begin{matrix}
 1&0\\
 0&0
    \end{matrix}\right)$ of the former choice,
    \begin{align}
        \rho_j(g_\phi).\left(\begin{matrix}
 1&0\\
 0&0
    \end{matrix}\right). \rho_l(g_\phi^{-1})=\left(\begin{matrix}
\cos j \phi \cos l \phi& \cos j \phi \sin l \phi\\
  \cos l \phi  \sin j \phi & \sin j \phi \sin l \phi
    \end{matrix}\right)
    \end{align}
Doing the same for the other elements of the canonical basis we arrive to a basis of solutions of \eqref{steerabilityconstraintjl}:
\begin{equation}\label{SO(2)solutionjl}
  \mathcal{B}^{(j,l)}_{\text{SO}(2)}=   \left\{ \begin{matrix}
         
     \left(\begin{matrix}
        \cos j\phi \cos l\phi &  \cos j\phi \sin l\phi\\
        \sin j\phi \cos l\phi & \sin j\phi \sin l\phi
\end{matrix}\right),\left(\begin{matrix}
       - \cos j\phi \sin l\phi &  \cos j\phi \cos l\phi\\
       - \sin j\phi \sin l\phi & \sin j\phi \cos l\phi
    \end{matrix}\right),\\
    \left(\begin{matrix}
        -\sin j\phi \cos l\phi & - \sin j\phi \sin l\phi\\
        \cos j\phi \cos l\phi & \cos j\phi \sin l\phi
\end{matrix}\right),\left(\begin{matrix}
        \sin j\phi \sin l\phi &  -\sin j\phi \cos l\phi\\
        -\cos j\phi \sin l\phi & \cos j\phi \cos l\phi
    \end{matrix}\right)
    \end{matrix}\right\}
\end{equation}
\end{enumerate}
As opposed to \cite{weiler2019E2CNN}, we did not need to pick from the start  a particular basis of functions, like the circular harmonics. These are given by how we write the representations $\rho_l$ and $\rho_j$.   
\subsubsection*{Comparing complex and real representations}

The complex basis \eqref{SO(2)solutionljcomplex} is one-dimensional, so as a real vector space is two-dimensional. However, the basis  for the real case \eqref{SO(2)solutionjl} is 4-dimensional, which seems like a contradiction. As noticed in \cite{weiler2019E2CNN},  the answer comes from the fact that the (inequivalent) real  representations \eqref{SO(2)irreps} are labeled by $j \geq 0$, roughly half the represenations in the complex case which are labeled by $j \in \mathbb{Z}$.   More precisely, there is an underlying real structure on the direct sum of complex unirreps $\rho_j \oplus \rho_{-j}$. So we need both $j$ and $-j$ to get one real unirrep \eqref{SO(2)irreps}.

\subsection{O$(2)$}

Now, we have 2D rotations and also a reflection $r_y:(x,y)\mapsto (x,-y)$. The other reflection $r_x: (x,y)\mapsto (-x,y)$ is a combination of  a $\pi$ rotation and $r_y$. Any group element is parameterized by an angle $\phi$ and the sign $s=\pm 1$,
\begin{equation}\label{SO(2)irreps}
     g_{\phi,s}=\left(\begin{matrix}
    \cos \phi & -s\sin \phi\\
    \sin  \phi & s\cos \phi
\end{matrix}\right)
 \end{equation}
 This is equivalent to saying that the following relation holds, 
 \begin{equation}\label{O(2)relation}
     r_y g_\phi r_y =g_{-\phi}, \qquad g_\phi\in\text{SO}(2).
\end{equation}
The $G-$orbits are the same as before since reflections leave the circle invariant. Picking $x_0=(R,0)^T$, we see that the stabilizer group $H$ is now given by 
\begin{equation}
    H=\left\{\mathbb{I}_{2\times 2},r_y\right\},\qquad r_y=\left(\begin{matrix}
    1 & 0\\
    0  & -1
\end{matrix}\right)
\end{equation}

\subsubsection*{Complex representations}

The complex unirreps of O$(2)$ differ substantially from those of SO$(2)$. This is because of the relation \eqref{O(2)relation}, which cannot be satisfied by the 1-dimensional unirreps of SO$(2)$ \eqref{SO(2)unirrepscomplex}. Instead, we have two 1-dimensional representations and infinite 2-dimensional ones labeled by $n\geq 1$,
\begin{equation}\label{O(2)irrepscomplex}
     \rho_0(g_{\phi,s})=1,\quad \rho_{\tilde 0}(g_{\phi,s})=s,\quad\rho_n(g_{\phi,1})=\left(\begin{matrix}
    e^{in\phi} & 0\\
    0 & e^{-in\phi}
\end{matrix}\right),\quad\rho_n(r_y)=\left(\begin{matrix}
    0 & 1\\
    1 & 0
\end{matrix}\right),\qquad n\geq 1
\end{equation}
Since considering the complex representations does not seem to simplify the procedure, compared to the real representations which are the ones we are really interested in, we stop here and move to the real case.

\subsubsection*{Real representations}
Irreducible representations are again of dimension one and two, but now \textit{we have two} 1-dimensional representations: 
\begin{equation}\label{SO(2)irreps}
     \rho_0(g_{\phi,s})=1,\quad \rho_{\tilde 0}(g_{\phi,s})=s,\quad\rho_j(g_{\phi,s})=\left(\begin{matrix}
    \cos j\phi & -s\sin j\phi\\
    \sin j \phi & s\cos j\phi
\end{matrix}\right)
 \end{equation}

We mimic what we did for SO$(2)$ and pick $x_0=(R,0)$, but now the stabilizer has an additional element: $r_y=g_{0,-}$. Then, 

The restricted representations are
\begin{equation}
    \begin{matrix}
 \rho^H_0(\mathbb{I}_{2\times 2})=1,& \rho^H_{\tilde 0}(\mathbb{I}_{2\times 2})=1,&
    \rho_j^H(\mathbb{I}_{2\times 2})=\mathbb{I}_{2\times 2}\\
    & & \\
    \rho^H_0(r_y)=1,& \rho^H_{\tilde 0}(r_y)=-1,& 
    \rho_j^H(r_y)=r_y
    \end{matrix}
    \end{equation}
The consistency constraint \eqref{constraintisotropy} has to be satisfied for both elements, of which we already analyzed the identity when studying SO$(2)$ and obtained a basis of solutions. More over, the new one-dimensional representation $\rho_{\tilde 0}$ also gives $1$ on the identity matrix (as it should) so does not impose anything new. Therefore in what follows we analyze the other element $r_y$, which may give additional constraints to be satisfied:  
\begin{enumerate}
    \item[$\mathbf{(0,0)}$:] $K^{(0,0)}$ is a real number and since $\rho^H_0(r_y)=1$ there is nothing else to impose. Also steering does not do anything since $\rho_0$ is the trivial representation.
    \item[$\mathbf{(0,\tilde 0)}$:] Here \eqref{constraintisotropy} when evaluated on $r_y$ reads $K^{(0,\tilde 0)}(x_0)=1.K^{(0,\tilde 0)}(x_0).(-1)$, which has no non-zero solution. 
    \item[$\mathbf{(\tilde 0,\tilde 0)}$:] Here \eqref{constraintisotropy} when evaluated on $r_y$ reads $K^{(\tilde 0,\tilde 0)}(x_0)=(-1).K^{(\tilde 0,\tilde 0)}(x_0).(-1)$. So there is nothing else to impose and steering does nothing.
    \item[$\mathbf{(0,l)}$:] In this case  \eqref{constraintisotropy}, when evaluated on $r_y$, reads $K^{(0,l)}(x_0)=1.K^{(0,l)}(x_0).r_y$, which only $e_1=(1,0)$ solves it.  
    We have then, \begin{equation}\label{O(2)solution0l}
    \mathcal{B}^{(0,l)}_{\text{O}(2)}=\left\{ (\cos l\phi,\, \sin l\phi)\right\}
\end{equation}
In Table 9 of \cite{weiler2019E2CNN} this solution is assigned to the following case, and viceversa. 
    \item[$\mathbf{(\tilde 0,l)}$:] Similarly to the previous case,  \eqref{constraintisotropy} when evaluated on $r_y$ now reads $K^{(\tilde 0,l)}(x_0)=(-1).K^{(
    \tilde 0,l)}(x_0).r_y$, which only $e_2=(0,1)$ solves it.  
    We have then, \begin{equation}\label{O(2)solution0l}
    \mathcal{B}^{(\tilde 0,l)}_{\text{O}(2)}=\left\{ (-\sin l\phi,\, \cos l\phi)\right\}
\end{equation}
 \item[$\mathbf{(j,l)}$:] In this case $K^{(j,l)}(x_0)$ is a 2-by-2 matrix, and \eqref{constraintisotropy} with $h=r_y$ reads $K^{(j,l)}(x_0)=r_y.K^{(j,l)}(x_0).r_y$. Considering the canonical basis, it is only solved by matrices with a 1 in the 11 and 22 sites (which are diagonal and so commute with $r_y$). In terms of the other basis using Pauli matrices discussed for SO$(2)$, it is solved by the identity and $\sigma_3$ (since $r_y=\sigma_3$). If we choose the canonical basis and steer again the matrix $\left(\begin{matrix}
 1&0\\
 0&0
    \end{matrix}\right)$,
    \begin{align}
        \rho_j(g_{\phi,s}).\left(\begin{matrix}
 1&0\\
 0&0
    \end{matrix}\right). \rho_l(g_{\phi,s}^{-1})=\left(\begin{matrix}
\cos j \phi \cos l \phi& \cos j \phi \sin l \phi\\
  \cos l \phi  \sin j \phi & \sin j \phi \sin l \phi
    \end{matrix}\right)
    \end{align}

Doing the same for the other diagonal element of the canonical basis we arrive to a basis of solutions of \eqref{steerabilityconstraintjl}:
\begin{equation}\label{O(2)solutionjl}
  \mathcal{B}^{(j,l)}_{\text{O}(2)}=   \left\{ \begin{matrix}
         
     \left(\begin{matrix}
        \cos j\phi \cos l\phi &  \cos j\phi \sin l\phi\\
        \sin j\phi \cos l\phi & \sin j\phi \sin l\phi
\end{matrix}\right),\left(\begin{matrix}
        \sin j\phi \sin l\phi &  -\sin j\phi \cos l\phi\\
        -\cos j\phi \sin l\phi & \cos j\phi \cos l\phi
    \end{matrix}\right)
    \end{matrix}\right\}
\end{equation}
Steering the identity and $\sigma_3$ gives precisely the basis of the $(j,l)$ case in Table 9 of \cite{weiler2019E2CNN}. 
\end{enumerate}

\subsection{SO$(3)$}
This is the group of 3D rotations of $\mathbb{R}^3$, and its orbits are the origin and spheres. Let us again pick some arbitrary radius $R$ and care only about the angles on the sphere $S^2$. Let us pick $x_0$ as the North Pole $x_0=(0,0,R)$. The stabilizer $H$ is then the group of rotations around the $z$ axis since they leave $x_0$ invariant, that is $H\simeq $ SO$(2)$, with
\begin{equation}
    H=\left\{g_\theta:=\left(\begin{matrix}
        R(\theta) &0\\
        0 & 1
    \end{matrix}\right);\quad R(\theta)=\left(\begin{matrix}
        \cos\theta & -\sin\theta\\
        \sin\theta &\cos\theta
    \end{matrix}\right),\quad \theta \in [0,2\pi)\right\}
\end{equation}

\subsubsection*{Complex representations}

The complex unirreps of SO$(3)$ are labeled by an integer $l\geq 0$ and the corresponding vector spaces $V_l$ have dimension $2l+1$. So we have only one unirrep for each (odd) dimension.  In terms of the Euler angles\footnote{The idea is to describe any 3D rotation as  a composition of rotations $z-y-z$. So $g_{\alpha,\beta,\gamma}=e^{-i\alpha L_z} e^{-i\beta L_y} e^{-i\gamma L_z}$ .} we have
\begin{equation}\label{SO(3)irrepscomplex}
    \rho_l(g_{\alpha,\beta,\gamma})=D^l(\alpha,\beta,\gamma),\qquad D^l_{mm'}(\alpha,\beta,\gamma):=e^{-im\alpha}d_{mm'}^l(\beta)e^{-im'\gamma} 
\end{equation}
with respect to a basis of spherical harmonics $Y^l_m \in L^2(S^2)$,
\begin{equation}
    g_{\alpha,\beta,\gamma}\cdot Y^l_m = Y^l_m \circ g_{\alpha,\beta,\gamma}^{-1}=\sum_{m'=-l}^{l} D_{m'm}^{l}(g_{\alpha,\beta,\gamma}) Y^l_{m'},
\end{equation}
where $m\in (-l,l)$ is an integer. Here $d_{mm'}^l(\beta)$ are the matrix elements of a rotation of angle $\beta$ around the $y$ axes.  

The restriction of these representations to the isotropy subgroup $H$ is
\begin{equation}\label{rhoHso(2)complex}
    \rho_l^H(g_{\theta})=\text{diag}\left(e^{-il\theta },e^{-i(l-1)\theta }, \dots, e^{il\theta }\right)=\bigoplus_{m=-l}^l e^{-im\theta }
\end{equation}
and the blocks in the diagonal are the complex unirreps of SO$(2)$. This is precisely the decomposition \eqref{decomposition}. Following the discussion around that equation, we can consider \eqref{constraintisotropy}, 
\begin{equation}
    K^{(j,l)}(x_0)=\rho_j^H(g_\theta).K^{(j,l)}(x_0).\rho_l^H(g_{-\theta})
\end{equation}
for a generic case $(j,l)$, and write $K^{(j,l)}(x_0)$ as 
\begin{equation}\label{complexsumofSO(2)}
    K^{(j,l)}(x_0)=\bigoplus_{m_l\in[-l,l],\,m_j \in[-j,j]} K^{(j,l)}_{m_l,m_j}(x_0).
\end{equation}
Notice that there are no multiplicities in \eqref{rhoHso(2)complex}, each irreducible representation appears only once. By Schur's Lemma, all these $K^{(j,l)}_{m_l,m_j}$ are zeros when $m_l\neq m_j$, and complex numbers if $m_l=m_j$,
\begin{equation}\label{complexSO(2)solution}
    K^{(j,l)}_{m_l,m_j}(x_0)\propto
    \delta_{m_l,m_j} 
\end{equation}
In order to see how this looks like more clearly, let 
\begin{equation}
    |l\,\,m\rangle :=Y^l_m
\end{equation}
denote a spherical harmonic more abstractly, using the bra-ket notation of quantum mechanics. Then $\langle l\,\,m|$ is the corresponding dual element, that is
\begin{equation}
    \langle l'\,\, m'|l\,\,m\rangle=\delta_{ll'} \delta_{mm'}
\end{equation}
With this notation, let us consider the matrix $T^{(j,l)}_{m}=| j \,\,m\rangle\langle l\,\,m|\in\mathbb{C}^{(2j+1) \times (2l+1)}$ which is made of zeros except at the intersection of the $\langle j \,\,m|$ row and $|l\,\,m\rangle $ column where it is 1, and such that,
\begin{equation}\label{SO(3)weightscondition}
-\text{min} (j,l)\leq m\leq \text{min}(j,l).    
\end{equation}
 So we see $T^{(j,l)}_{m}$ maps the $m$-th irrep in the decomposition of $\rho_l^H$ to the same $m$-th irrep in the decomposition of $\rho_j^H$.   When we steer it with \eqref{steerabilityconstraintjl}, using \eqref{SO(3)irrepscomplex}, we obtain the matrix elements of $K^{(j,l)}_{m,m}(g_{\alpha,\beta,\gamma}\cdot x_0)$ up to a multiplicative complex number,
\begin{equation}\label{SO(3)steercomplex}
    \langle j\,\,m_j|K^{(j,l)}_{m,m}(g_{\alpha,\beta,\gamma}\cdot x_0)|l\,\,m_l\rangle\propto D^j_{m_jm}(g_{\alpha,\beta,\gamma})D^l_{mm_l}(g_{\alpha,\beta,\gamma})^{-1}
\end{equation}
So we found a complete basis of complex solutions to \eqref{steerabilityconstraintjl},
\begin{align}\label{SO(3)solutionjlcomplex}
    \mathcal{B}^{(j,l)}_{\text{SO}(3)}=&\left\{M^{(j,l)}_{m}\in L^2(S^2,\mathbb{C}^{(2j+1) \times (2l+1)})\,\,;\right.\nonumber\\
    &\quad  \langle j\,\, m_j|M^{(j,l)}_m|l\,\,m_l\rangle=D^j_{m_jm}(g_{\alpha,\beta,\gamma})D^l_{mm_l}(g_{\alpha,\beta,\gamma})^{-1},\nonumber\\
    &\quad -\text{min} (j,l)\leq m\leq \text{min}(j,l) \Bigr\}\qquad (\text{complex representations})
\end{align}

A few comments are in order:
\begin{enumerate}
\item We have arrived to a closed formula for the matrix elements of the intertwiners, which is ready to use. 

\item More over, the information needed to construct this basis is minimal: the matrix elements of the Wigner D-matrices which appear in the steerability condition \eqref{steerabilityconstraintjl}.

\item It may seem that these intertwiners $M_m^{(j,l)}$ depend on three angles $\alpha$, $\beta$ and $\gamma$, instead of being functions on a two-sphere, as they should. It turns out that they do not depend on $\gamma$, since in the product of elements of the Wigner D-matrices the $\gamma$ dependence can be seen to disappear by definition of the matrices.  

    \item Notice that compared to \cite{weiler20183d}, we do not need to change to another basis where the tensor product  $\rho_j \otimes \rho_l$ decomposes as direct sum of irreps (the so called coupled basis), and then change back to the original uncoupled basis at the end. This change of basis required knowing the matrix $Q$ that implements the change of basis, and $Q$ can be written with Clebsch-Gordan coefficients or found by a numerical approach as in \cite{weiler20183d}. In the basis we use, all these steps are unnecessary, and at the end of the day one arrives to the same vector space of solutions, but written with different basis elements. 
\end{enumerate}

\subsubsection*{Real representations}

The real unirreps are obtained from the complex ones by changing the basis from the spherical harmonics to the real spherical harmonics,
\begin{align}
    Y^{c}_{lm}&=\frac{1}{\sqrt{2}}\left(Y^l_m + (-1)^m Y^l_{-m}\right)\\
    Y^{s}_{lm}&=\frac{-i}{\sqrt{2}}\left(Y^l_m - (-1)^m Y^l_{-m}\right)\\
    Y^{R}_{l0}&=Y^{l}_{0}
\end{align}
with $m\in [1,l]$. Let us denote by $S^l$ the matrix that implements this change from the spherical harmonics $Y^l_m$ to the real ones $Y_{lm}$. We shall employ the ket notation for the real spherical harmonics as, 
\begin{equation}\label{realharmonicskets}
    |l\,\,m\rangle_{\text{real}}:=\left\{ \begin{matrix}
        Y^c_{lm},& m>0\\
        Y^s_{lm},& m<0\\
        Y_{l0},& m=0
        
    \end{matrix}\right.
\end{equation}
The real version $R^j$ of the Wigner D-matrices $D^j$ is then 
\begin{equation}\label{realR}
    R^j=(S^j)^*D^j(S^j)^{T},\qquad 
    R^j_{m'm}(g):={}_{\text{real}}\langle j\,\, m'|R^j(g)|j\,\, m\rangle_{\text{real}}
\end{equation}
Knowing the Wigner D-matrices and  $S^j$ we can compute the real matrices $R^j$ and store them once and for all. 

The restriction of the real representation to $H$ is
\begin{equation}\label{rhoHSO(2)}
    \rho_l^H(g_{\theta})=\text{diag}\left(1,R(\theta),R(\theta)^2, \dots, R(\theta)^l\right)
\end{equation}
written in a basis of real spherical harmonics of the form
\begin{equation}\label{realsphericalharmonics}
(Y_{l0}^R,Y_{l1}^c,Y_{l1}^s,\dots,Y_{ll}^c,Y_{ll}^s)^T = S^l  (Y_{ll},\dots,Y_{l0},\dots,Y_{l-l})^T   
\end{equation}
We can decompose $\rho_l^H$ in a direct sum of the irreducible one and two-dimensional real representations of SO$(2)$. Actually  \eqref{rhoHSO(2)} is already in block-diagonal form.  The difference with the complex case is evident: then we had only 1-dimensional irreps, now we have always one 1-dimensional irrep and the others are all 2-dimensional.

Let us repeat what we did before with the complex representation: the stabilizer constraint  \eqref{constraintisotropy} 
reads 
\begin{equation}
    K^{(j,l)}(x_0)=\rho_j^H(g_\theta).K^{(j,l)}(x_0).\rho_l^H(g_{-\theta})
\end{equation}
This means that $K^{(j,l)}$ intertwines between the reducible representations $\rho_l^H$ and $\rho_j^H$. As we already discussed for the complex case,  $K^{(j,l)}(x_0)$ can be written as a block matrix where each block is an equivariant map between irreducible representations of SO$(2)$, now the real ones,
\begin{equation}\label{sumofSO(2)}
    K^{(j,l)}(x_0)=\bigoplus_{m_l\in[0,l],\,m_j \in[0,j]} K^{(j,l)}_{m_l,m_j}(x_0) 
\end{equation}
where $m_l=0$ is the trivial representation and the others are two-dimensional irreducible representations indicating the power of $R(\theta)$ in \eqref{rhoHSO(2)}, and the same for $m_j$. Notice once again that there are no multiplicities in \eqref{rhoHSO(2)}. Therefore we apply Schur's Lemma to each block $K^{(j,l)}_{m_l,m_j}(x_0)$,
\begin{equation}\label{realSO(2)solution}
    K^{(j,l)}_{m_l,m_j}(x_0)\propto \left\{ 
    \begin{matrix}
    1,& m_l=m_j=0\\
    \delta_{m_l,m_j} (a\mathbb{I}_{2\times 2}+b J),
    & \text{otherwise}\end{matrix}\right.
\end{equation}
with $a,b\in\mathbb{R}$ and $J$ defined as
\begin{equation}
    J=\left(\begin{matrix}
        0&-1\\ 1&0
    \end{matrix}\right).
\end{equation}
Again, this forces $0\leq m_l=m_j=m$ satisfying also \eqref{SO(3)weightscondition}, so we have a $2 \text{min}(j,l)+1$-dimensional space of independent real intertwiners. 
As opposed to the complex case, we have to be careful in applying Schur's Lemma: the \textit{real} two-dimensional representations are interwined by ``a scalar'' times the identity but this scalar belongs to  End${}_{\text{SO}(2)}(\mathbb{R}^2)$, which is a vector space generated by $\mathbb{I}_{2\times 2}$ and $J$. 

Finally we have to steer each $K^{(j,l)}_{m,m}(x_0)$, 
\begin{equation}
    K^{(j,l)}(g\cdot x_0)=\rho_j(g) \left( \bigoplus_{m\in[0,\text{min}(l,j)]}K^{(j,l)}_{m,m}(x_0)\right) \rho_l(g^{-1})
\end{equation}
We can steer each $K^{(j,l)}_{m,m}(x_0)$ and sum. This is actually a nice way to see that a possible basis of solutions to \eqref{steerabilityconstraintjl} is given by steering simpler elements as $K^{(j,l)}_{m,m}(x_0)$. Let us separate between three cases: $m=0$, $m >0$ with $b=0$ and $m >0$ with $a=0$. In the first case we proceed analogously to the complex case and using \eqref{realR} arrive at, 
\begin{align}\label{SO(3)steer00}
    \left(K^{(j,l)}_{0,0}(g_{\alpha,\beta,\gamma}\cdot x_0)\right)_{m_jm_l}&\propto R^j_{m_j0}(g_{\alpha,\beta,\gamma})R^l_{0m_l}(g^{-1}_{\alpha,\beta,\gamma})\nonumber\\ &=\left((S^j)^*D^j(g_{\alpha,\beta,\gamma})(S^j)^{T}\right)_{m_j0}\left((S^l)^*D^l(g_{\alpha,\beta,\gamma})^{-1}(S^l)^{T}\right)_{0m_l}
\end{align}
in the real basis \eqref{realsphericalharmonics}.

For the case $m > 0$ and $b=0$, which corresponds to the 2-dimensional identity in the $m$-th subspace, we can write the corresponding intertwiner $T^{(j,l)}_{(m,\mathbb{I})}$ as, 
\begin{equation}
    T^{(j,l)}_{(m,\mathbb{I})}={}_{\text{real}}|j\,\,m\rangle\langle l\,\, m|_{\text{real}} +{}_{\text{real}}|j\,\,-m\rangle\langle l\,\, -m|_{\text{real}}
\end{equation}
which when steered gives,
\begin{equation}\label{SO(3)steermI}
    \left(K^{(j,l)}_{m,m,\mathbb{I}}(g_{\alpha,\beta,\gamma}\cdot x_0)\right)_{m_jm_l}\propto R^j_{m_j,m}(g_{\alpha,\beta,\gamma})R^l_{m,m_l}(g^{-1}_{\alpha,\beta,\gamma})+R^j_{m_j,-m}(g_{\alpha,\beta,\gamma})R^l_{-m,m_l}(g^{-1}_{\alpha,\beta,\gamma}).
\end{equation}
Similarly, for the case $m > 0$ and $a=0$, we can write 
\begin{equation}
    T^{(j,l)}_{(m,J)}={}_{\text{real}}|j\,\,-m\rangle\langle l\,\, m|_{\text{real}} -{}_{\text{real}}|j\,\,m\rangle\langle l\,\, -m|_{\text{real}}
\end{equation}
which when steered gives,
\begin{equation}\label{SO(3)steermJ}
    \left(K^{(j,l)}_{m,m,J}(g_{\alpha,\beta,\gamma}\cdot x_0)\right)_{m_jm_l}\propto R^j_{m_j,-m}(g_{\alpha,\beta,\gamma})R^l_{m,m_l}(g^{-1}_{\alpha,\beta,\gamma})-R^j_{m_j,m}(g_{\alpha,\beta,\gamma})R^l_{-m,m_l}(g^{-1}_{\alpha,\beta,\gamma}).
\end{equation}
This can be expressed in terms of the Wigner D-matrices using \eqref{realR} as in \eqref{SO(3)steer00}.

In summary, we have found a basis of real solutions to \eqref{steerabilityconstraintjl},
\begin{align}\label{SO(3)solutionjl}
    \mathcal{B}^{(j,l)}_{\text{SO}(3)}=&\left\{M^{(j,l)}_{0,0},M^{(j,l)}_{m,\mathbb{I}},M^{(j,l)}_{m,J}\in L^2(S^2,\mathbb{R}^{(2j+1) \times (2l+1)})\,\,;\right.\nonumber\\
    &\quad{}_{\text{real}}\langle j\,\, m_j|M^{(j,l)}|l\,\,m_l\rangle_{\text{real}}=\text{RHS of }\eqref{SO(3)steer00},\eqref{SO(3)steermI},\eqref{SO(3)steermJ}\,\,\text{respectively, with} \,\nonumber\\
    &\quad 0< m\leq \text{min}(j,l) \Bigr\}
\end{align}
The comments made after \eqref{SO(3)solutionjlcomplex} are relevant also for this real case. 

\subsubsection*{Comparing complex and real representations}

In the complex case, we have found a basis made of 2min$ (j,l)+  1$  complex-linear intertwiners, so we have a (4min$ (j,l)+  2$)-dimensional space of real learnable parameters. On the contrary, in the real case we have 2min$ (j,l)+  1$ real-linear intertwiners and therefore a (2min$ (j,l)+  1$)-dimensional space of real learnable parameters. To connect the solutions to the complex case with those of the real case, we can argue as follows: take the complex SO$(2)$-intertwiners $T_m^{(j,l)}=|j\,\, m\rangle \langle l\,\, m|$, combine them to give the general solution 
\[K^{(j,l)}(x_0)=\sum_{m=-l}^l c_m T_m^{(j,l)},\] and demand that they commute with the (anti-linear) canonical conjugation 
\[C_l:|l\,\, m\rangle  \mapsto (-1)^m |l\,\, -m\rangle .\]
This readily implies that 
\begin{equation}
    c_0\in\mathbb{R}, \quad c_m^*=c_{-m},\quad m>0
\end{equation}
and we get the reduction by half of the real learnable parameters. More over, writing $c_m=a_n+ib_m$, and transforming the complex bases $\langle l\,\,m|$ and $|j\,\,m\rangle$ with $(S^l)^{-1}$ and $S^j$ respectively, we get that for a given $m>0$, 
\begin{align}
    c_m |j\,\,m\rangle \langle l\,\,m| +
    c_{-m} |j\,\,-m\rangle \langle l\,\,-m|&=a_m\left(
    {}_{\text{real}}|j\,\,m\rangle \langle l\,\,m|_{\text{real}} +{}_{\text{real}}|j\,\,-m\rangle \langle l\,\,-m|_{\text{real}}\right)\nonumber\\
    &+b_m \left(
    {}_{\text{real}}|j\,\,-m\rangle \langle l\,\,m|_{\text{real}} -{}_{\text{real}}|j\,\,m\rangle \langle l\,\,-m|_{\text{real}}\right)\nonumber\\
    &=a_m \mathbb{I}_{2\times 2}+b_m J,
\end{align}
so we end up with the basis of real-linear SO$(2)$-intertwiners we found using the real representations.  
\subsection{O$(3)$}

This is the group of matrices leaving the Euclidean distance to the origin invariant. It can be thought as the group of 3D rotations together with reflections, which have determinant $-1$. In particular $-\mathbb{I} \in$ O$(3)$, but does not belong to SO$(3)$. Here $\mathbb{I}=\mathbb{I}_{3\times 3}$ is the 3-dimensional identity matrix and we will just call it $\mathbb{I}$ from now on. Taking $\mathbb{Z}_2$ the group generated by $-\mathbb{I}$, we have an isomorphism
\begin{equation}
    \varphi:(A,Z)\mapsto AZ, \quad A\in\text{SO}(3),\,Z=\pm \mathbb{I}
\end{equation}
between SO$(3)\times \mathbb{Z}_2$  and O$(3)$.

Since reflections leave the spheres invariant, spheres are again the orbits of the group together with the origin. Let us consider again the North pole $x_0=(0,0,R)$, with isotropy group $H$ given by the rotations around the $z$ axis and also reflections with normal perpendicular to the $z$ axis, as $r_y=\text{diag}(1,-1,1)$. This means  $H\simeq $ O$(2)$, with
\begin{equation}
    H=\left\{g_{\theta,s}:=\left(\begin{matrix}
        R_s(\theta) &0\\
        0 & 1
    \end{matrix}\right);\quad R_s(\theta)=\left(\begin{matrix}
        \cos\theta & -s\sin\theta\\
        \sin\theta &s\cos\theta
    \end{matrix}\right),\quad \theta \in [0,2\pi),\quad s=\pm 1\right\}
\end{equation}
Here $R_1(\theta)$ is a rotation in the $x-y$ plane, and $R_{-1}(\theta)=R_1(\theta/2)\sigma_3 R_1(-\theta/2)$ represents a reflection with a reflection plane whose normal direction is given by $\theta/2$ measured from the $y$ 
axes. 

\subsubsection*{Complex representations}

Using the fact that O$(3)$ is (isomorphic to) the direct product SO$(3)\times \mathbb{Z}_2$, we can use the irreps of SO$(3)$ made of spherical harmonics $Y^l_m$, and see how an element like $-\mathbb{I}\in$ O$(3)/$SO$(3)$ acts on $Y^l_m$. It is well-known that under parity\footnote{Parity means a change in sign of the cartesian coordinates $(x_1,x_2,x_3)\in\mathbb{R}^3$, namely an inversion.} the spherical harmonics change as $Y^l_m\mapsto (-1)^l Y^l_m$, so we have two inequivalent representations\footnote{Because of the fact that $\rho_l$ restricted to SO$(3)$ is irreducible, by Schur's Lemma $\rho_l(-\mathbb{I})$ has to be $\pm 1$ on each $l$-space.},
\begin{equation}\label{O(3)irrepcomplex}
    \rho_{l,\pm}(-\mathbb{I})=\pm (-1)^l.
\end{equation}
The plus sign corresponds to the change of $Y^l_m$ under parity and the minus sign to the the same tensored with the determinant representation.

The restriction of these representations to the isotropy subgroup $H\simeq$ O$(2)$ can be understood as follows: since $g_{\theta,s}$ can be decomposed as a rotation around $z$ of angle $\theta$ and a reflection $r_y$ in the $x-z$ plane, we have that 
\begin{equation}
    \rho^H_{l,\pm}|(g_{\theta,s})=\rho_l(g_{\theta,1})\rho_{l,\pm}(r_y)^{(1-s)/2}
\end{equation}
where $\rho_l(g_{\theta,1})$ is just $e^{-im\theta}\delta_{m,m'}$ in the basis of spherical harmonics with $-l\leq m\leq l$.  So we need to compute $\rho_{l,\pm}(r_y)$ and for this we use the fact that $r_y=-R_y(\pi)$, namely minus a rotation around $y$ of angle $\pi$,
\begin{equation}
   (\rho_{l,\pm}(r_y))_{m,m'}=\pm(-1)^l d_{m,m'}^l(\pi)=\pm(-1)^m \delta_{m,-m'}.  
\end{equation}
Equivalently, on the 2-dimensional subspaces spanned by $Y^l_m$ and $Y^l_{-m}$ with $m>0$, 
\begin{equation}
     \rho_{l,\pm}(g_{\theta,1})=\pm\left(\begin{matrix}
         e^{-im\theta}&0\\0&e^{im\theta}
     \end{matrix}\right),\qquad \rho_{l,\pm}(r_y)=\pm (-1)^m\left(\begin{matrix}
         0&1\\1&0
     \end{matrix}\right).
\end{equation}
Then, calling $\rho_{m,\pm}$ these \textit{equivalent}  O$(2)$ irreps, 
\begin{equation}\label{rhoHo(2)complex}
    \rho_{l,\pm}^H(g_{\theta,s})=\pm\rho_0 \oplus\bigoplus_{m=1}^l \rho_{m,\pm}\qquad \text{(complex representation)}
\end{equation}
where $\rho_0$ is the trivial representation on the 1-dimensional subspace spanned by $Y^l_0$. We have once again arrived at the decomposition \eqref{decomposition}, now for isotropy subgroup O$(2)$. It is somewhat similar to the real representation of SO$(3)$ restricted to  $H\simeq$ SO$(2)$, having a 1-dimensional representation and then 2-dimensional representations. However, here we have a complex representation, so Schur's Lemma says that the space of intertwiners is isomorphic the the complex numbers. More explicitly,
\begin{equation}\label{complexsumofO(2)}
    K^{((j,\epsilon_j),(l,\epsilon_l))}(x_0)=\bigoplus_{m_l,m_j\geq 0} K^{((j,\epsilon_j),(l,\epsilon_l))}_{m_l,m_j}(x_0),\qquad \epsilon_j,\epsilon_l\in\left\{+,-\right\}
\end{equation}
with 
\begin{equation}\label{O(2)solutioncomplex}
    K^{((j,\epsilon_j),(l,\epsilon_l))}_{m_l,m_j}(x_0)\propto \left\{ 
    \begin{matrix}
    \delta_{\epsilon_l,\epsilon_j},& m_l=m_j=0\\
    \delta_{m_l,m_j}(\delta_{\epsilon_l,\epsilon_j} \mathbb{I}_{2\times 2}+\delta_{\epsilon_l,-\epsilon_j}\sigma_3),
    & \text{otherwise}\end{matrix}\right.
\end{equation}
We see that there can be steerable kernels between different sign choices, as is immediate to check that $\sigma_3$ intertwines between $\rho_{m,+}$ and $\rho_{m,-}$.  
For each pair of O$(3)$ irreps, there are min$(l,j)+1$ complex intertwiners.  When we steer them with \eqref{steerabilityconstraintjl}, using \eqref{SO(3)irrepscomplex} and \eqref{O(3)irrepcomplex},
\begin{align}\label{O(3)steercomplex}
    \langle j\,\,m_j|K^{((j,\epsilon),(l,\pm\epsilon))}_{m,m}(g_{\alpha,\beta,\gamma}\cdot x_0)|l\,\,m_l\rangle &\propto D^j_{m_jm}(g_{\alpha,\beta,\gamma})D^l_{mm_l}(g_{\alpha,\beta,\gamma})^{-1}\nonumber\\&\pm D^j_{m_j,-m}(g_{\alpha,\beta,\gamma})D^l_{-m,m_l}(g_{\alpha,\beta,\gamma})^{-1},\quad \epsilon=\pm.
\end{align} 
So, we have
\begin{align}\label{O(3)solutionjlcomplex}
    \mathcal{B}^{((j,\epsilon),(l,\pm\epsilon))}_{\text{O}(3)}=&\left\{M^{(j,l)}_{m}\in L^2(S^2,\mathbb{C}^{(2j+1) \times (2l+1)})\,\,;\right.\nonumber\\
    &\quad  \langle j\,\, m_j|M^{(j,l)}_m|l\,\,m_l\rangle=D^j_{m_jm}(g_{\alpha,\beta,\gamma})D^l_{mm_l}(g_{\alpha,\beta,\gamma})^{-1}\nonumber\\
    &\qquad \qquad\qquad \qquad \quad\,\pm D^j_{m_j,-m}(g_{\alpha,\beta,\gamma})D^l_{-m,m_l}(g_{\alpha,\beta,\gamma})^{-1},\nonumber\\
    &\quad 0\leq m\leq \text{min}(j,l)\Bigr\}\qquad (\text{complex representations})
\end{align}
Here we end up with less and different intertwiners compared to SO$(3)$ (notice that $m$ needs to be non-negative), for each sign choice (see \eqref{SO(3)solutionjlcomplex}). More concretely, me have specific linear combinations of the intertwiners of \eqref{SO(3)solutionjlcomplex}.

\subsubsection*{Real representations}

For the real representations, we can argue similarly to what we did for the real representations of SO$(3)$. Notice that we have to change to the real spherical harmonics basis \eqref{realsphericalharmonics} the complex solutions \eqref{O(2)solutioncomplex} which are essentially given by $\mathbb{I}_{2\times 2}$ and $\sigma_3$ in the space spanned by $Y^l_m$ and $Y^l_{-m}$. So changing basis the identity gives the identity once again, and changing basis $\sigma_3$ gives $-iJ=-\sigma_2$. This means that we need a real coefficient for $\epsilon_l=\epsilon_j$ and a purely imaginary coefficient for $\epsilon_l=-\epsilon_j$ if we want to restrict the complex solutions \eqref{O(2)solutioncomplex} to the real case. In the end, we arrive at,
\begin{align}\label{O(3)solutionjl}
    \mathcal{B}^{((j,\epsilon),(l,\pm\epsilon))}_{\text{O}(3)}=&\left\{M^{(j,l)}_{m}\in L^2(S^2,\mathbb{R}^{(2j+1) \times (2l+1)})\,\,;\right.\nonumber\\
    &\quad  {}_{\text{real}}\langle j\,\, m_j|M^{(j,l)}_m|l\,\,m_l\rangle_{\text{real}}=R^j_{m_jm}(g_{\alpha,\beta,\gamma})R^l_{\pm  m,m_l}(g_{\alpha,\beta,\gamma})^{-1}\nonumber\\
    &\qquad \qquad\qquad \qquad \quad\,\pm R^j_{m_j,-m}(g_{\alpha,\beta,\gamma})R^l_{\mp m,m_l}(g_{\alpha,\beta,\gamma})^{-1},\nonumber\\
    &\quad 0\leq m\leq \text{min}(j,l)\Bigr\}\qquad (\text{real representations})
\end{align}
using the notation defined in  \eqref{realharmonicskets}.

\subsection{The Lorentz group SO$(1,3)$}

We will consider now the proper orthochronous Lorentz group SO${}^+(1,3)$, given by those $4\times 4$ real matrices $\Lambda$ that preserve the Minkowski product \begin{equation}\label{Minkowskiproduct}
    ||x||^2:=x^\mu x^\nu \eta_{\mu\nu}=(x^0)^2-|\vec{x}|^2
\end{equation} and preserve the direction of time and orientation (have unit determinant). We are considering $\mu=0,1,2,3$ and $\eta=$diag$(1,-1,-1,-1)$ as is standard in the particle physics community. Latin indexes such as $i,j,k$ range from $1$ to $3$. Note that $\eta^{-1}$ has the same diagonal form as $\eta$ and has contravariant indices, namely its components are $\eta^{\mu\nu}$, such that $\eta_{\mu\alpha}\eta^{\alpha\nu}=\delta^\nu_\mu$. A generic Lorentz transformation $\Lambda$ is made from three independent rotations (like the Euler angles used before)  and three real parameters which describe boosts in each possible spatial direction (change of inertial observers).

The real irreducible representations of this group come from the complex representations of its double cover SL$(2,\mathbb{C})$. Notice that SO${}^+(1,3)$ is not compact, so its finite irreducible representations which are of interest to us cannot be unitary. We give now a short summary of the finite complex and real irreps of SO${}^+(1,3)$, based mainly on \cite{sternberg1995group} and \cite{knapp2001representation}.

First, let us establish an isomorphism of metric spaces between Minkowski $\mathbb{R}^{1,3}=(\mathbb{R}^{4},\eta)$ and $(u(2),\text{det})$ which is the space of Hermitian $2\times 2 $ matrices with metric given by the determinant. Here the metrics are pseudo-Riemannian, allowing vectors of non-positive norm. The relation is constructed using the Pauli matrices and defining $\sigma_\mu=(\mathbb{I}_{2\times 2},\vec{\sigma})$. Then a general element of $u(2)$ can be written as $X:=x^\mu \sigma_\mu$. The isomorphism $\Phi:\mathbb{R}^{1,3}\rightarrow u(2)$ is,
 \begin{equation}\label{Phi}
 \Phi(x)=x^\mu \sigma_\mu=X    
 \end{equation}
It is a simple computation to check that $x^\mu x^\nu \eta_{\mu\nu}=\text{det}(X)$. So we can view Minkowski space-time as $u(2)$.

Second, we can show there is a clear relation between SL$(2,\mathbb{C})$ and SO${}^+(1,3)$ by considering the action of SL$(2,\mathbb{C})$ on $u(2)$,
\begin{equation}\label{SL2Caction}
    g\cdot X=gXg^{\dagger},\qquad g\in \text{SL}(2,\mathbb{C}).
\end{equation}
This action preserves the determinant det$(X)$, since $g$ has unit determinant, which means it preserves $\eta$ on Minkowski space-time. Therefore, this action is actually a group homomorphism from SL$(2,\mathbb{C})$ to SO${}^+(1,3)$. Since $g$ and $-g$ give the same action, it turns out that SO${}^+(1,3)=$SL$(2,\mathbb{C})/\mathbb{Z}_2$. Explicitly, 
\begin{equation}\label{Lambdafromg}
    g\mapsto \phi(g)= \Lambda^\mu_{\,\,\nu}=\frac{1}{2}\delta^{\mu\rho}\text{Tr}(\sigma_\rho g\sigma_\nu g^\dagger).
\end{equation}
Actually, we have explained so far that $\Lambda$ is an element of O$(1,3)$, but since SL$(2,\mathbb{C})$ is connected it turns out that the image of  $\phi$ is SO${}^+(1,3)$. 

 Third, by a result called the unitary trick, the holomorphic finite-dimensional representations
of SL$(2,\mathbb{C})$ are in one-to-one correspondence with the complex finite-dimensional representations of SU$(2)$. In the end, one needs to complexify the groups, and ends up with the action
\begin{equation}\label{SL2Ccomplexactions}
    (g_L,g_R) \cdot X:=g_L X g_R^\dagger,\qquad (g_L,g_R)\in \text{SL}(2,\mathbb{C})\times\text{SL}(2,\mathbb{C}). 
\end{equation}
Then $(g,g)\in \text{SL}(2,\mathbb{C})\times\text{SL}(2,\mathbb{C})$ represents a Lorentz transformation by \eqref{SL2Caction} and what followed. Because of the unitary trick, we can then describe the finite-dimensional irreps of $\text{SL}(2,\mathbb{C})\times\text{SL}(2,\mathbb{C})$ by two unitary finite-dimensional irreps of SU$(2)$, and thus they are labeled by  a pair of half-integers $(j_L,j_R)$. 

Concretely, an irrep $\rho_{(j_L,j_R)}$ of $\text{SL}(2,\mathbb{C})\times\text{SL}(2,\mathbb{C})$
 acts on the space of homogeneous polynomials of variables $z_1,z_2,\bar{z}_1,\bar{z}_2$, of degree $2j_L$ on the variables $z_1,z_2$, and degree $2j_R$  on the variables $\bar{z}_1,\bar{z}_2$, which can be seen\footnote{The second copy of SL$(2,\mathbb{C})$ acts with the conjugate representation, which comes from \eqref{SL2Ccomplexactions}. The conjugate to the standard  representation is usually denoted with a dot $v_{\dot a}\mapsto \bar{g}_{\dot a \dot b}v_{\dot  b}$ in the physics literature, but we shall not adopt this here to avoid cluttering. We just need to remember the first sentence of this footnote.} as $\text{Sym}^{2j_L}\mathbb{C}^{2} \otimes \text{Sym}^{2j_R}\bar{\mathbb{C}}^{2}$. If $P$ is such a polynomial, then,
 \begin{equation}
     \rho_{(j_L,j_R)}(g)\,P=P\circ g^{-1}.
 \end{equation}
Real finite-dimensional representations of the (proper orthochronous) Lorentz group are the ones for which $j_L=j_R$ or direct sums\footnote{These are physically relevant since for example Dirac fermions such as the electron transform as $(1/2,0)\oplus (0,1/2)$.} of the form $(j_L,j_R) \oplus (j_R,j_L) $. The former are irreducible, and reality comes from the fact that these representations are the ones invariant under complex conjugation given by the exchange $j_L \leftrightarrow j_R$. We will only consider (real) intertwiners between these real representations.

\subsubsection{Massive particles}

From \eqref{Minkowskiproduct} it can be seen that the orbits are the hypersurfaces in $\mathbb{R}^{1,3}$ which preserve that product. We will first consider (one sheet of) the so-called massive hyperboloids, which are those for which $x_0>0$ and $||x||^2=m^2>0$. We pick a point on it, say $(m,0,0,0)$ with $m>0$, and its stabilizer is clearly given by all 3D rotations and reflections which act on the space components $\vec{x}$. So $H\simeq$ SO$(3)$.

Recall that SO$(3)$ is the image of SU$(2)$ under $\phi$ in \eqref{Lambdafromg}. Then, we have that  $\rho_{(j_L,j_R)}^H$ is the tensor product of two SU$(2)$ irreps, and we can use the well-known Clebsch-Gordan coefficients of SU$(2)$ to decompose it in irreps:
\begin{equation}\label{rhoLorentzSU(2)}
   \mathbb{C}^{2j_L+1} \otimes \mathbb{C}^{2j_R+1}\simeq\bigoplus_{B=|j_L-j_R|}^{j_L+j_R} \mathbb{C}^{2B+1} \overset{\scriptscriptstyle  j_L=j_R=j}{=} \bigoplus_{B=0}^{2j} \mathbb{C}^{2B+1} 
\end{equation}
In order to keep expressions simple, we keep writing $\mathbb{C}^{2B+1}$ as the representation vector space although in what follows we will make sure that representations and intertwiners are real-linear. For example the real structure behind $\mathbb{C}^2$ is $\mathbb{R}^4$ and in general the real structure of $\mathbb{C}^{2B+1}$ is $\mathbb{R}^{4B+2}$.

For steerable kernels beween $(l,l)$ and $(j,j)$ representations, we match each invariant subspace $\mathbb{C}^{2A+1}$ and $\mathbb{C}^{2B+1} $  in the above decomposition ($0\leq A \leq 2l$). A basis of real intertwiners are of the form
\begin{equation} \label{LorentzT}
    T_{(A,B)}^{(j,l)}=\delta_{A,B} \times \left\{\begin{matrix}
        \mathbb{I}, & A=B\quad  \text{integers}\\
      a  \,\mathbb{I} + b\, I + c\,J+d\, K,&\qquad A=B \quad \text{half-integers}
    \end{matrix}\right.
\end{equation}
 where $\mathbb{I}$ is the identity map, $I$ is multiplication by $i$, $J$ is charge conjugation, $K=IJ$, and the coefficients are all real. This gives rise to a quaternionic structure. However we do not need it at this point, since from \eqref{rhoLorentzSU(2)} we see that the irreducible subspaces correspond to integer $A$ and $B$. 

 For steerable kernels where one or both of the (real) representations are of the form $(j_L,j_R) \oplus (j_R,j_L)$, we have 
 \begin{equation}\label{rhoLorentzSU(2)directsum}
   \mathbb{C}^{(2j_L+1)\times (2j_R+1) }\, \oplus\,  \mathbb{C}^{(2j_R+1)\times (2j_L+1) }\simeq\bigoplus_{B_1=|j_L-j_R|}^{j_L+j_R} \mathbb{C}^{2B_1+1} \oplus \bigoplus_{B_2=|j_L-j_R|}^{j_L+j_R} \mathbb{C}^{2B_2+1} = \bigoplus_{B=|j_L-j_R|}^{j_L+j_R} \left(\mathbb{C}^{2B+1}\right)^{\oplus 2} 
\end{equation}
so we see that each irrep appears with multiplicity 2. We also note that in this case there could be half-integers $A$ or $B$, so quaternionic-like intertwiners may appear as in \eqref{LorentzT}. 

In any case, the point we need to stress is that one has to match the same irreducible subspaces, namely $A=B$, and if one or both  have multiplicity 2, there is an internal matrix with real or quaternionic entries. Let us be more explicit
\begin{enumerate}
    \item $\mathbf{l=(l_L,l_R)\oplus(l_R,l_L)\, \text{  and  } j=(j_L,j_L)\text{:} }$ From now on we call $j=j_L$. As we saw, $B$ are non-negative integers up to $2j$. And $A \in [\,|l_L-l_R|,\, l_L+l_R \,]$. It is clear that if $A$ runs on half-integers, there is no intertwiner. The irreducible spaces  $\mathbb{C}^{2A+1}$ come with multiplicity 2, so if there is a match $A=B$, the intertwiner between  $\mathbb{C}^{2A+1}$  and $\mathbb{C}^{2B+1}$ can take two independent real values, so it is a matrix $\mathbb{R}^{1\times  2}\otimes\mathbb{I}$. Clearly if we switch input and output representations, the intertwiner between two spaces labeled by $A=B$ is a matrix $\mathbb{R}^{2\times 1}\otimes \mathbb{I}$.
\item $\mathbf{l=(l_L,l_R)\oplus(l_R,l_L) \text{  and  } j=(l_L,j_R)\oplus(j_R,j_L)\text{:}}$ Now both $A$ and $B$ label irreps of multiplicity 2 and which may be integers or half-integers. Say there is a match $A=B$. If they are integers, we have an intertwiner of the form $\mathbb{R}^{2\times 2}\otimes \mathbb{I}$. If they are both half-integers, the intertwiner is  a matrix $\mathbb{H}^{2\times 2}\otimes \mathbb{I}$, where $\mathbb{H}$ stands for the quaternionic space\footnote{For the intertwiner of spin $1/2$ spaces, which are of real dimension 4, we have:
\[I=
\begin{pmatrix}
0 & -1 & 0 & 0 \\
1 &  0 & 0 & 0 \\
0 &  0 & 0 & -1 \\
0 &  0 & 1 &  0
\end{pmatrix},\quad J =
\begin{pmatrix}
0 & 0 & -1 & 0 \\
0 & 0 &  0 & 1 \\
1 & 0 &  0 & 0 \\
0 & -1 & 0 & 0
\end{pmatrix},\quad K =
\begin{pmatrix}
0 & 0 & 0 & -1 \\
0 & 0 & -1 &  0 \\
0 & 1 & 0 &  0 \\
1 & 0 & 0 &  0
\end{pmatrix}\].} spanned by $\mathbb{I}$, $I$, $J$ and $K$ as explained after \eqref{LorentzT}.   
\end{enumerate}

Once we have the intertwiners between some particular irreducible spaces, we have to steer them to obtain $K^{((j_L,j_R),(l_L,l_R))}(g\cdot x_0)$ according to \eqref{steerabilityconstraintjl}, as we did before for the compact groups. If $A=B=L$, $L$ an integer, then we need to compute
\begin{equation}
    \rho_{(j_L,j_R)}(g)T_{LL}\rho_{(l_L,l_R)}(g^{-1})=\rho_{(j_L,j_R)}(g) \left(\sum_{m=-L}^L |L\,\,m\rangle\otimes \langle L\,\,m|\right) \rho_{(l_L,l_R)}(g^{-1})
\end{equation}
where $|L\,\,m\rangle$ form the coupled basis of $\mathbb{C}^{2L+1}$. We can rewrite this in terms of the input and output original bases using\footnote{One would have to take care of the fact that the ``right'' copy is acted upon the conjugate representation, which is isomorphic to the fundamental one.} SU$(2)$ Clebsch-Gordan coefficients (and \textit{not the Lorentz ones}),
\begin{align}
    \rho_{(j_L,j_R)}(g)T_{LL}\rho_{(l_L,l_R)}(g^{-1})&=\rho_{(j_L,j_R)}(g) \left(\sum_{m=-L}^L\sum_{m_L=-j_L}^{j_L}\sum_{m_R=-j_R}^{j_R}  \langle j_L,j_R;m_{L},m_R|L\,\,m\rangle |j_L,j_R;m_{L},m_R\rangle\right.\nonumber\\
    &\left.\otimes \sum_{m'_L=-l_L}^{l_L}\sum_{m'_R=-l_R}^{l_R} \langle L\,\,m| l_L,l_R;m'_{L},m'_R\rangle \langle l_L,l_R;m'_{L},m'_R|\right) \rho_{(l_L,l_R)}(g^{-1})
\end{align}
where $\langle j_L,j_R;m_{L},m_R|L\,\,m\rangle $ are the C-G coefficients needed to pass from the uncoupled basis labeled by $(m_L,m_R)$ to the coupled one labeled by $(L,m)$. However, it may be more useful for practical purposes to work with the standard representation of SO$(3,1)$, that is 4-vectors such as\footnote{We are setting the speed of light to 1, that is $c=1$.} $x^\mu=(t,\vec{x})$, and its tensor products. 

Let us do this in detail for the $(1/2,1/2) $ representation which is equivalent to the space of 4-vectors. To any element of the uncoupled basis $\left\{|m_L,m_R\rangle\right\}$ we can assign a 2-by-2 Hermitian matrix by , 
\begin{equation}\label{Twospins4vector}
    |m_L,m_R\rangle\mapsto X_{m_L,m_R}:=\frac{1}{2}(M_{m_L,m_R   }+(M_{m_L,m_R   })^\dagger)
\end{equation}
where $M_{m_L,m_R}$ is a matrix where the only non-zero entry is 1 at the $(m_L,m_R)$ position. The linear combination at the RHS of \eqref{Twospins4vector}, $X_{m_L,m_R}$, gives a Hermitian matrix (which span a real 4-dimensional vector space). And we know from the discussion at the beginning of this Subsection that 2-by-2 Hermitian matrices are in 1-to-1 correspondence with 4-vectors by \eqref{Phi}. The uncoupled basis $\left\{|m_L,m_R\rangle\right\}$ is actually a basis for a complex 4-dimensional space, but the subspace invariant under the conjugation that exchanges $m_L$ with  $m_R$ is a real 4-dimensional space, isomorphic to the space spanned by $X_{m_L,m_R}$. Concretely, we have the following identifications between elements of bases of real 4-dimensional vector spaces,
\begin{align}
    |++\rangle + |--\rangle \leftrightarrow&\,\,  \mathbb{I} \leftrightarrow e_0  \nonumber\\
    |+-\rangle + |-+\rangle \leftrightarrow&\,\, \sigma_1 \leftrightarrow e_1  \nonumber\\
-i|+-\rangle + i|-+\rangle \leftrightarrow&\,\, \sigma_2 \leftrightarrow e_2  \nonumber\\ 
|++\rangle - |--\rangle \leftrightarrow&\,\, \sigma_3 \leftrightarrow e_3  
\end{align}
where $\left\{e_\mu\right\}$ are the canonical 4-vectors. A general 4-vector $V$ is then $V=V^\mu e_\mu$. A $(p,0)$ tensor can be formed by linear combinations of tensor products of $p$ 4-vectors, so a good basis of $(p,0)$-tensors is $\left\{e_{\mu_1}\otimes\dots\otimes e_{\mu_p}\right \}$. There are also dual elements $\left\{\tilde{e}^\nu\right\}$ to the 4-vectors $\left\{e_\mu\right\}$, such that $\tilde{e}^\nu(e_\mu)=\delta^\nu_\mu$. A Lorentz group matrix $\Lambda$ acts on them with the inverse element $\Lambda^{-1}$. Then we have \textit{covariant} $(0,q)$-tensors given by linear combinations of a basis such as  $\left\{\tilde e^{\mu_1}\otimes\dots\otimes \tilde e^{\mu_q}\right \}$. Finally we have $(p,q)$-tensors given by linear combinations of the tensor product basis $\left\{e_{\mu_1}\otimes\dots\otimes e_{\mu_p}\otimes \tilde e^{\mu_1}\otimes\dots\otimes \tilde e^{\mu_q}\right \}$. 

It turns out that under SO$(3)$, 4-vectors are a direct sum of spin 0 (spanned by $e_0$) and spin 1 objects (spanned by $e_i$), while a $(2,0)$-tensor $Q$ is a direct sum of the scalar $Q^{00}$ (spin 0), two spin 1 vectors $Q^{0i}$ and $Q^{i0}$, and a rank 2 3D tensor $Q^{ij}$ which itself splits into another scalar $\sum_{i=1}^3 Q^{ii}$, a vector given by its anti-symmetric part $\frac12 (Q^{ij}-Q^{ji})$ and a spin 2 tensor given by its symmetric traceless part $\frac12 (Q^{ij}+Q^{ji})-\frac13 \delta^{ij} \sum_{i=k}^3 Q^{kk}$. Hopefully it is clear that the important point is to understand the steering of intertwiners between the irreducible parts under SO$(3)$ of these kind of spacetime tensors. We do this for a few cases now, which should serve as guide for more general spins sitting inside arbitrary spacetime tensors. We will see that since SO$(3)$ intertwiners are proportional to the identity, sitting inside a spacetime tensor representation they look as projectors.

\subsubsection*{$\textbf{A=B=0}$}

Let us begin with the simplest case, where one or both of the input and output representations contain the spin 0 representation (we omit from now on the labels of the input and output representations) . This representation is contained in the 4-vector representation discussed above, and is given by the subspace spanned by $e_0$ since it is the fixed point of SO$(3)$ rotations (any other $e_i$ rotates). Being 0 an integer, the intertwiner is proportional to the 1-dimensional identity, as already discussed. So inside the standard representation of 4-vectors,
\begin{equation}
    T_{(0,0)}=e_0\otimes \tilde e^0
\end{equation}
It is clear that it acts as the identity on $e_0$ and annihilates the other three 4-vectors $e_i$.
In spacetime components it reads 
\begin{equation}
    (T_{(0)})^\mu{}_\nu = \delta^\mu_0 \delta^0_\nu .
\end{equation}
Now we steer $T_{(0)}$ by a Lorentz group element $\Lambda$ and readily obtain a basis of intertwiners made of a single element,
\begin{equation}
\mathcal{B}^{(\text{spin }0)}_{\text{SO}^+(1,3)}=\left\{(K_0(\Lambda))^\mu{}_\nu = u^\mu u_\nu\right\} .
\end{equation}
where 
\begin{equation}
    u^\mu := \Lambda^\mu_{\,\,\nu}(e_0)^\nu=\Lambda^\mu_{\,\,0},\qquad u_\mu :=\eta_{\mu\nu}u^{\nu}=(\Lambda^{-1})^0_{\,\,\mu}. 
\end{equation}
This means that $u$ is still a unit timelike vector (as it should).

\subsubsection*{$\textbf{A=B=1}$}

Similarly to the previous case, we know the intertwiner is proportional to the identity of the three-dimensional space spanned by $\left\{e_i\right\}$.  If this spin 1 subspace lives inside a 4-vector space, 
\begin{equation}
    T_{(1,1)}=e_\mu\otimes \tilde e^\mu-e_0\otimes \tilde e^0,
\end{equation}
which in spacetime components reads,
\begin{equation}
  (T_{(1,1)})^\mu{}_\nu
= \delta^\mu{}_\nu - \delta^\mu_0 \delta^0_\nu .
\end{equation}
We now steer it,
\begin{equation}
\mathcal{B}^{(\text{spin }1 \,\subset\, 4-\text{Vector})}_{\text{SO}^+(1,3)}=\left\{
(K_{(1)}(\Lambda))^\mu{}_\nu
= \delta^\mu{}_\nu - u^\mu u_\nu=:\Delta^\mu{}_\nu\right\}.
\end{equation}
Notice that $\Delta^\mu{}_\nu$ satisfies $\Delta^\mu{}_\nu \Delta^\nu{}_\sigma=\Delta^\mu{}_\sigma$ and annihilates $u$, meanining that it projects onto the space orthogonal to $u$.

If it is the case that the spin 1 we are interested in lives inside a $(2,0)$ tensor $Q=Q^{\mu\nu} e_\mu\otimes e_\nu$, it can be of the form\footnote{We have also $Q_{\bar 1}:=Q^{0i} e_0 \otimes e_i$.} $Q_{1}:=Q^{i0} e_i \otimes e_0$ or the antisymmetric part of $Q^{ij}$, namely $Q_{AS}:=\frac12 (Q^{ij}-Q^{ji})\,e_i \otimes e_j$. For the former,
\begin{equation}
    T_{(1,1)}^{Q_1}=(e_\mu\otimes \tilde e^\mu-e_0\otimes \tilde e^0)\otimes (e_0\otimes \tilde e^0) 
\end{equation}
This gives the 3-dimensional identity in the first factor and the 1-dimensional identity in the second factor. In components reads,
\begin{equation}\label{vectorin2tensor}
    (T_{(1,1)}^{Q_1})^{\mu\nu}_{\,\,\,\,\rho\sigma}=(\delta^\mu_\rho  -\delta^\mu_0 \delta^0_\rho) \delta^\nu_0 \delta^0_\sigma
\end{equation}
For the latter, 
\begin{equation}
    T_{(1,1)}^{Q_{AS}}=\frac12 \left((e_i\otimes \tilde e^i)\otimes  (e_j\otimes \tilde e^j)-(e_i\otimes \tilde e^j)\otimes  (e_j\otimes \tilde e^i) \right)
\end{equation}
which is the projector onto the anti-symmetric spatial part of the $(2,0)$-tensor: the first term is a spatial identity operator and the second term interchanges the spatial basis vectors. In components we get,
\begin{equation}\label{3DASpart2tensor}
    (T_{(1,1)}^{Q_{AS}})^{\mu\nu}_{\,\,\,\,\rho\sigma}=\frac12 \left((\delta^\mu_\rho  -\delta^\mu_0 \delta^0_\rho)(\delta^\nu_\sigma  -\delta^\nu_0 \delta^0_\sigma )-(\delta^\nu_\rho  -\delta^\nu_0 \delta^0_\rho)(\delta^\mu_\sigma  -\delta^\mu_0 \delta^0_\sigma) \right)
\end{equation}
Finally we steer \eqref{vectorin2tensor} and \eqref{3DASpart2tensor},
\begin{align}\label{spin1in2tensorsteered}
    \mathcal{B}^{(Q_1 \,\subset\, (2,0)-\text{Tensor})}_{\text{SO}^+(1,3)}&=\left\{(K_{(1,1)}^{Q_1}(\Lambda))^{\mu\nu}_{\,\,\,\,\rho\sigma}=\Delta^\mu_{\,\,\rho}  u^\nu u_\sigma\right\}\nonumber\\
  \mathcal{B}^{(Q_{\text{AS}}\,\subset\, (2,0)-\text{Tensor})}_{\text{SO}^+(1,3)}&=\left\{ (K_{(1,1)}^{Q_{AS}}(\Lambda))^{\mu\nu}_{\,\,\,\,\rho\sigma}=\frac12 \left(\Delta^\mu_{\,\,\rho}\Delta^\nu_{\,\,\sigma}-\Delta^\nu_{\,\,\rho}\Delta^\mu_{\,\,\sigma}\right)\right\}
\end{align}

\subsubsection*{$\textbf{A=B=2}$}

Here we consider a spatial rank-2 tensor that sits inside a $(2,0)$ tensor, that is the traceless symmetric part of $Q^{ij}$, $Q_S:=\frac12 (Q^{ij}+Q^{ji}-\frac13 \delta^{ij} \text{ Tr }Q)\,e_i \otimes e_j$. Once again, having integer spin 2, the intertwiner is proportional to the 5-dimensional identity of this space of traceless symmetric spatial tensors. However inside the space of $(2,0)$ tensors we should view it as a projector $Q \mapsto Q_S$, which is similar to the anti-symmetric case but this time is the identity \textit{plus} the interchange of spatical basis vectors. Then, 
\begin{equation}
 T_{(2,2)}^{Q_{S}}=\frac12 \left((e_i\otimes \tilde e^i)\otimes  (e_j\otimes \tilde e^j)+(e_i\otimes \tilde e^j)\otimes  (e_j\otimes \tilde e^i) \right)-\frac13 (e_i\otimes \tilde e^j) \otimes (e_i\otimes \tilde e^j)
\end{equation}
where we subtracted the trace. In components,
\begin{equation}\label{3DSpart2tensor}
    (T_{(2,2)}^{Q_{S}})^{\mu\nu}_{\,\,\,\,\rho\sigma}=\frac12 \left((\delta^\mu_\rho  -\delta^\mu_0 \delta^0_\rho)(\delta^\nu_\sigma  -\delta^\nu_0 \delta^0_\sigma )+(\delta^\nu_\rho  -\delta^\nu_0 \delta^0_\rho)(\delta^\mu_\sigma  -\delta^\mu_0 \delta^0_\sigma) \right)-\frac13 (\eta^{\mu\nu}-\delta_0^\mu\delta_0^\nu)(\eta_{\rho\sigma}-\delta_\rho^0\delta_\sigma^0)
\end{equation}
When we steer it,
\begin{equation}
   \mathcal{B}^{(Q_{\text{S}}\,\subset\, (2,0)-\text{Tensor})}_{\text{SO}^+(1,3)}=\left\{  (K_{(2,2)}(\Lambda))^{\mu\nu}{}_{\rho\sigma}
=
\frac12\Big(
\Delta^\mu{}_\rho\Delta^\nu{}_\sigma
+\Delta^\mu{}_\sigma\Delta^\nu{}_\rho
\Big)
-\frac{1}{3}\Delta^{\mu\nu}\Delta_{\rho\sigma}\right\}.
\end{equation}
where,
\begin{equation}
    \Delta^{\mu\nu}:=\eta^{\nu\alpha}\Delta^\mu{}_\alpha,\qquad \Delta_{\mu\nu}:=\eta_{\mu\alpha}\Delta^\alpha{}_\nu.
\end{equation}

\subsubsection*{Comment: intertwiners as projectors}

Let us emphasize a point we have mentioned already: an integer-spin intertwiner is nothing but a projector onto the space of the corresponding invariant subspace, sitting inside a spacetime tensor representation. For example the spin-0 intertwiner  $e_0 \otimes \tilde e^0 = \mathbb{I}_{1\times 1}\oplus 0_{3\times 3}$ projects onto 
the subspace spanned by $e_0=(1,0,0,0)^\mu e_\mu$, inside the 4-vector representation, and when we steer with $\Lambda$ we can interpret it as changing to a different inertial reference frame where the unit timelike vector is now  $u = u^\mu e_\mu$, so the projector reads $u^\mu u_\nu$. Analogously, for example the spin-1 intertwiner for tensors of the form $Q^{i0}$ inside a $(2,0)$-tensor is given by $(e_\mu\otimes \tilde e^\mu-e_0\otimes \tilde e^0)\otimes (e_0\otimes \tilde e^0) = (0_{1\times 1}\oplus \mathbb{I}_{3\times 3})\otimes  (\mathbb{I}_{1\times 1}\oplus 0_{3\times 3})$, which projects onto the spatial part the first index and onto the timelike direction the second index. When we steer we are again changing reference frame and writing covariantly the same projector:  $(K_{(1,1)}^{Q_1}(\Lambda))^{\mu\nu}_{\,\,\,\,\rho\sigma}=\Delta^\mu_{\,\,\rho}  u^\nu u_\sigma$. Ho

\subsubsection*{Some generalities of half-integer spins}

Here we give a short list of usefull formulae for half-integer spins, and then consider the cases of spin $1/2$ and $3/2$. Recall the defining relation of the $\gamma$ matrices,
\begin{equation}
    \left\{\gamma^\mu,\gamma^\nu\right\}=2\eta^{\mu\nu}.
\end{equation}
which entails a matrix representation of the Clifford algebra. 
Let us define the following quantities,
\begin{equation}
\slashed{u}:=u_\mu\gamma^\mu,\qquad
\gamma_\perp^\mu := \Delta^\mu{}_\nu\,\gamma^\nu,
\end{equation}
which immediately imply,
\begin{equation}
\slashed{u}^2=1,\qquad
u_\mu\gamma_\perp^\mu=0.    
\end{equation}

As previously mentioned, we need the operation of charge conjugation, which can be defined as
\begin{equation}
  \mathcal C(\psi) := C\gamma^{0}\,{\psi}^*,
\end{equation}
with 
\begin{equation}
C^{-1} \gamma^\mu C=-(\gamma^\mu)^T    
\end{equation}
for some fixed reference frame where $\gamma^0$ and $\gamma^i$ are chosen Hermitian and anti-Hermitian respectively\footnote{When boosted, the $\gamma$ matrices transform such that hermiticity conditions fail.} and for $\gamma^0$ real or purely imaginary. These conditions hold both for the Dirac basis as well as the Weyl basis and Majorana basis. From this considerations, if $S(\Lambda)$ is a spinorial representation of the Lorentz group, it can be shown that 
\begin{equation}
    S(\Lambda)\,\mathcal C\,S(\Lambda)^{-1} = \mathcal C 
\end{equation}
meaning that complex conjugation is invariant under Lorentz transformations. This tells us that when we steer, the intertwiner made from $\mathcal{C}$ will remain unchanged! Actually, the following four intertwiners remain unchanged, 
\begin{equation}
    \mathbb I,\qquad I:= i\,\mathbb I,\qquad J:=\mathcal C,\qquad K:= i\,\mathcal C .
\end{equation}

\subsubsection*{$\mathbf{A=B=1/2}$} For spin $1/2$ we have,
\begin{equation}
    T_{1/2}(u)= a\,\mathbb I + b\,i\mathbb I + c\,\mathcal C + d\,i\mathcal C,
\qquad a,b,c,d\in\mathbb R.
\end{equation}
which acts on the spinorial index. 

When we steer only the direction of propagation $e_0$ in the comoving frame changes, since as we already discussed, the spinorial part of the intertwiner remains unchanged. The positive/negative energy projectors with respect to $u$ read,
\begin{equation}
    P_\pm(u) := \frac12(1\pm \slashed{u}),
\end{equation}
So when we steer we arrive at
\begin{equation}
    \mathcal{B}^{(1/2)}_{\text{SO}^+(1,3)}=\left\{K_{(1/2)}(u)
=
\sum_{\epsilon=\pm}
\Big(
a_\epsilon\,\mathbb{I} + b_\epsilon\,i\mathbb{I} + c_\epsilon\,\mathcal C + d_\epsilon\,i\mathcal C
\Big)\,P_\epsilon(u)\right\}.
\end{equation}
with $a_\epsilon,b_\epsilon,c_\epsilon,d_\epsilon\in\mathbb R$.

\subsubsection*{$\mathbf{A=B=3/2}$} 
Now, spin $3/2$ typically arises as the irreducible subspace\footnote{This is known as a  Rarita-Schwinger field, however in order to isolate the spin $3/2$ one needs to impose the Dirac equation and the condition $\partial_\mu \psi^\mu=0$. Since we are not considering differential operators in this work, the intertwiners of this case should be taken as a first attempt and further investigation may be necessary. } of a spinor-vector $\psi^\mu$ such that $u_\mu\psi^\mu=\gamma_\mu\psi^\mu=0$ . It is a simple exercise to show that the following expression squares to itself acting on such a $\psi^\mu$ and anihhilates $u^\mu$, so it projects onto the subspace of $\psi$'s,
\begin{equation}
 \Pi_{3/2}(u)^{\mu}{}_{\nu}
:=
\Delta^{\mu}{}_{\nu}
-\frac13\,\gamma_\perp^{\mu}\gamma_{\perp\nu}.
\end{equation}
Similarly to the previous case, when steered,
\begin{equation}
    \mathcal{B}^{(3/2)}_{\text{SO}^+(1,3)}=\left\{(K_{(3/2)}(u))^\mu{}_\nu
=
\sum_{\epsilon=\pm}
\Big(
a_\epsilon\,\mathbb{I} + b_\epsilon\,i\mathbb{I} + c_\epsilon\,\mathcal C + d_\epsilon\,i\mathcal C
\Big)\,\Pi_{3/2}(u)^{\mu}{}_{\nu}\,P_\epsilon(u)\right\}.
\end{equation}

\subsubsection{Massless particles}

This case corresponds to the cone 
\begin{equation}
    \mathcal{P
    }=\left\{p\in\mathbb{R}^{1,3 };\quad p^2=0,\,\,p^0>0\right\}.
\end{equation}
We pick a point $n\in\mathcal{P}$, which defines a null direction. In some reference frame, $n^\mu=(1,0,0,1)$, so it represents light-speed motion  towards $z$.  It can be seen that the isotropy group is ISO$(2)$, meaning that it is generated by an SO$(2)$ coming from a rotation around $z$, and two ``translations'', which are not so direct to see (they are explained in detailed in  \cite{Weinberg_1995}). These translations imply the little group      is not semi-simple and therefore irreps do not restrict to completely reducible representations. Fortunately, in Nature we do not see particles that transform non-trivially under these translations, so we can forget about them and consider only the SO$(2)$ subgroup of the ISO$(2)$ little group\footnote{More precisely, we consider representations where the translation part acts trivially.}. 

Instead of studying the restricted irreps of Lorentz to this SO$(2)$ subgroup, we can take advantage of the lesson we learned from the massive case: we need projectors onto the invariant subspace where SO$(2)$ acts, roughly of the form $\mathbb{I} \oplus 0$.  Take $n$ and another null vector $\bar{n}$ such that 
\begin{equation}
    n\cdot \bar{ n}=1.
\end{equation}
Then a projector orthogonal to both $n$ and $\bar n$ is
\begin{equation}\label{transverseprojector}
    \Delta^{\mu\nu}=\eta^{\mu\nu}-n^\mu\bar{n}^\nu-n^\nu\bar{n}^\mu
\end{equation}
which transforms covariantly under Lorentz transformations. Contrary to the massive case, where the timelike unit vector $e_0$  has some unique role for a fixed reference frame representing a static observer, in the massless case there is no null vector with a preferred role with respect to the fixed reference frame. Changing reference frame makes no difference. What we can say is that $n$ is as good as null vector as any other $n'^\mu=\Lambda^\mu{}_\nu n^\nu$.    

There is a gauge choice in choosing $\bar{n}$, since we could also take
\begin{equation}
    \bar{n'}=\bar{n}+a_i e_i +\frac{1}{2} a^2 n
\end{equation}
where $e_i$ are two transverse 4-vectors, 
\begin{equation}
    e_i \cdot n=0,\quad e_i \cdot e_j=-\delta_{ij}, \quad a_i\in\mathbb{R},\quad a^2=a_ia_i.
\end{equation}
This is related to the translation part of the little group ISO$(2)$. If we choose $\bar{n'}$ instead of $\bar n$ to define \eqref{transverseprojector}, we get,
\begin{equation}
    \Delta'^{\mu\nu}=\Delta^{\mu\nu}-a_i(n^\mu e_i^\nu+n^\nu e_i^\mu)-a^2n^\mu n^\nu
\end{equation}

The point is that  the projector\eqref{transverseprojector} is the steered intertwiner we are looking for acting on the space of transverse 4-vectors \textit{up to addition of $n$}. This is expected, as the transverse space to $n$ is defined modulo addition of $n$: Consider for example a transverse wave propagating in the null direction $n$. The oscillations occur in the transverse directions $e_i$ (polarization vectors), but we could also take these directions to be $e_i+a_i n$. This ambiguity is intrinsic of the massless case but the task at hand, if physically sensible,  should not depend on the gauge choice. 

The basis of steerable kernels we have so far corresponds to a spin-1 field,
\begin{equation}\label{Lorentzmasslesssolutionspin1}
    \mathcal{B}^{(1)}_{\text{SO}^+(1,3)}=\left\{(K_{(1)}(n))^\mu{}_\nu
=\Delta^\mu{}
_\nu=\delta^{\mu}{}_\nu-n^\mu\bar{n}_\nu-n_\nu\bar{n}^\mu\right\},\quad \text(massless).
\end{equation}
This looks $n$-dependent but should be $\Lambda$-dependent.  If we steer we just make a Lorentz transformation, which changes $n$ and $\bar{n}$. However, we can first arbitrarily choose $n=n_0=(1,0,0,1)$, and then steer and get a generic $n^\mu=\Lambda^\mu{}_\nu  n_0^\nu=\Lambda^\mu{}_0+\Lambda^\mu{}_3$. So the steered final result is precisely   $\Delta^\mu{}
_\nu$ as in \eqref{transverseprojector} and \eqref{Lorentzmasslesssolutionspin1} is indeed the basis we are looking for where the information of the group element $\Lambda$ is encoded in $n^\mu=\Lambda^\mu{}_0+\Lambda^\mu{}_3$.  

For a spin-2 field, we can construct the projector  similarly to the symmetric traceless tensor of the massive case, and we arrive at,
\begin{equation}
    \mathcal{B}^{(2)}_{\text{SO}^+(1,3)}=\left\{(K_{(2)}(n))^{\mu\nu}{}_{\rho\sigma}
=\frac12 \left(\Delta^\mu_\rho\Delta^\nu_\sigma+\Delta^\mu_\sigma\Delta^\nu_\rho\right)-\frac12  \Delta^{\mu\nu}\Delta_{\rho\sigma}\right\},\quad \text(massless).
\end{equation}
Here we subtract the trace of the transverse space, which is 2-dimensional, that is why a factor $\frac12$ appears.

\section{Conclusions}
 
 In this work we have presented a general strategy for solving the kernel steerability constraint. This approach is more direct that the one in \cite{lang2021a} that uses Clebsch-Gordan coefficients in order to go to the coupled basis and back, and therefore saves on computational resources. More over,  hopefully this method is easier to understand for those not very familiar with techniques of representation theory. Also, it permits to obtain relatively easily explicit ready-to-use bases of steerable kernels for any compact Lie group. As such, it is possible to construct equivariant maps between arbitrary types of tensors without much considerable effort. 
 
 From  the implementation point of view, as usual aliasing needs to be addressed. Instead of placing a cut-off on the representation label $J$ of the coupled basis (which runs between some bounds set by $l$ and $j$), here we can place two independent cut-offs on the feature maps irreps $l$ and $j$, making  it more conceptually clear, more flexible and the network should become  more expressive. Of course, this ultimately needs to be shown in different experiments.

\section*{Acknowledgments}

I would like to thank Rodrigo Diaz for the invitation and encouragement to give a series of lectures at Instituto Tecnológico de Buenos Aires (ITBA) on equivariant CNNs, which triggered this project. This work was partially supported by grant PIP 11220210100685CO.

\bibliographystyle{toine}
\bibliography{biblio}{}
\end{document}